\documentclass[journal]{IEEEtran}

\usepackage[table]{xcolor}
\usepackage{tikz}
\usetikzlibrary{fit,shapes.geometric}

\newcounter{nodemarkers}

\usepackage{cite}
\usepackage{array,tabulary,tabularx,multirow}
\newcolumntype{Y}{>{\arraybackslash\centering}X}

\usepackage{multirow}

  \usepackage{graphicx}
  \usepackage{xcolor}
\usepackage[table]{xcolor}
\usepackage[table]{xcolor}

\usepackage{algorithmic}
\usepackage[ruled,vlined,linesnumbered]{algorithm2e}

\usepackage{amsmath}

\usepackage{algorithmic}
\usepackage{amsthm}

\usepackage[caption=false,font=footnotesize]{subfig}
\usepackage[flushleft]{threeparttable}

\usepackage{amssymb}

\DeclareMathOperator*{\argmax}{arg\,max}
\DeclareMathOperator*{\argmin}{arg\,min}
\newcommand\norm[1]{\left\lVert#1\right\rVert}
\def\BibTeX{{\rm B\kern-.05em{\sc i\kern-.025em b}\kern-.08em
    T\kern-.1667em\lower.7ex\hbox{E}\kern-.125emX}}
    
\begin{document}

\title{MAQ-CaF: A Modular Air Quality Calibration and Forecasting method for cross-sensitive  pollutants}
%
%
%

\author{Yousuf~Hashmy,~\IEEEmembership{Student Member,~IEEE,}
        Zill~Ullah~Khan,~\IEEEmembership{Student Member,~IEEE,}
        Rehan~Hafiz,~\IEEEmembership{Member,~IEEE}
        Usman~Younis,~\IEEEmembership{Senior Member,~IEEE}
        and~Tauseef~Tauqeer,~\IEEEmembership{Member,~IEEE}
\vspace{-8mm}

\thanks{Yousuf Hashmy, Zill Ullah Khan, Rehan Hafiz (rehan.hafiz@itu.edu.pk), Usman Younis and Tausif Tauqeer are with Information Technology University, Lahore, Pakistan.}
}

\maketitle


\begin{abstract}
The climatic challenges are rising across the globe in general and in worst hit under-developed countries in particular. The need for accurate measurements and forecasting of pollutants with low-cost deployment is more pertinent today than ever before. Low-cost air quality monitoring sensors are prone to erroneous measurements, frequent downtimes, and uncertain operational conditions. Such a situation demands a prudent approach to ensure an effective and flexible calibration scheme.
We propose MAQ-CaF, a modular air quality calibration, and forecasting methodology, that side-steps the challenges of unreliability through its modular machine learning-based design which leverages the potential of IoT framework. It stores the calibrated data both locally and remotely with an added feature of future predictions. Our specially designed validation process helps to establish the proposed solution's applicability and flexibility without compromising accuracy. $\textrm{CO}, \textrm{ SO}_2, \textrm{ NO}_2, \textrm{ O}_3, \textrm{ PM}_{1.0}, \textrm{ PM}_{2.5} \textrm{ and} \textrm{ PM}_{10}$ were calibrated and monitored with reasonable accuracy. Such an attempt is a step toward addressing climate change's global challenge through appropriate monitoring and air quality tracking across a wider geographical region via affordable monitoring.
\end{abstract}

\begin{IEEEkeywords}
Low cost sensors and devices, sensor calibration, cloud services and air quality.
\end{IEEEkeywords}

%
\IEEEpeerreviewmaketitle

\vspace{-4mm}
\section{Introduction}
\IEEEPARstart{G}{lobally}, around $90\%$ of the population breathe air that is non-compliant with WHO Air Quality Guidelines, resulting in a loss of around $3$ million human lives annually \cite{world2016ambient}. Moreover, degradation in air quality poses an imminent threat to work efficiency and economy as well, as indicated in \cite{jaimini2017investigation,ali2018pollution}. Less economically developed countries (LEDCs) are adopting low-cost sensing techniques to enhance spatio-temporal data density, without exceeding financial limitations \cite{munir2019analysing,kumar2019low}. One of the challenges faced by employing IoT-based low-cost sensor networks is the accuracy of measurements. Lately, the quality of information (QoI) metrics of such networks has transformed into an area of keen interest \cite{ferrer2019comparative}. Low-cost sensors are prone to erroneous readings, generally \cite{castell2017can,chiang2017design,mallires2019developing}.
\begin{figure}[t]
\centering
\includegraphics[width = 3.5in]{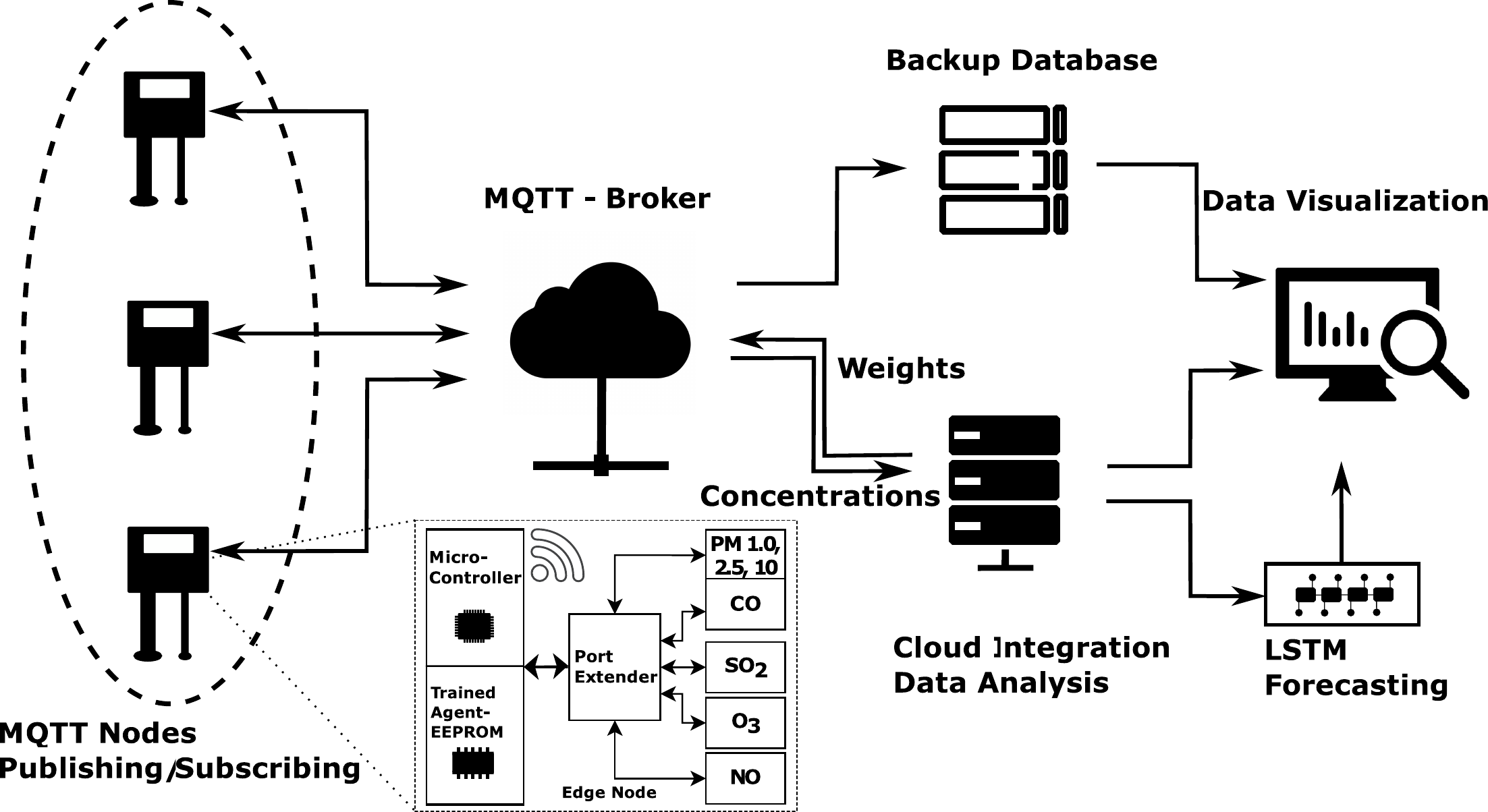}

\caption{The platform for monitoring and calibration of pollutant concentration on a large scale. Each sensor edge node consists of a micro-controller, a EEPROM, a port extender and air quality sensing elements. \vspace{-6mm}}
\label{iot} 
\end{figure}
To address the issues pertaining to the calibration of air quality metrics based upon IoT, researchers and scientists have been focusing on laboratory calibration \cite{castell2017can} or co-location calibration \cite{delaine2019situ,hagan2018calibration,sahu2020robust}. The former involves standard gas mixtures and superior-quality analytical instruments. Additionally, the non-linearities in the real measurements are not easily achieved in a controlled laboratory environment. The latter uses the references from some government-run systems to calibrate the low-cost sensors. The datasets are often made public such as EPA AirNow. In \cite{mueller2017design}, the co-located instruments have shown promise in calibrating $\textrm{NO}_2$ and $\textrm{O}_3$. 
Furthermore, \cite{lai2019iot} discusses the edge-computing infrastructure for the monitoring of air quality indicators. The details of cost comparisons of various platforms are provided in \cite{lai2019iot}, alluding to cost-effectiveness, and ease of implementation. The fusion of the data acquired from multi-sensor IoT platform for metal-oxide and electrochemical sensors for only two kinds of pollutants is described in \cite{ferrer2020multisensor,ha2020sensing}. \cite{ferrer2020multisensor} further indicates the need for the cross-sensitivity of the air quality indicators. Additionally, the work in \cite{idrees2020low}, discusses static as well as mobile air monitoring systems subject to their respective deployment strategies. The portal system requires larger logistics and increased costs. The inflexibility of the pollution monitoring and calibration schemes are also some of the major concerns \cite{idrees2020low}.

The measurements of the low precision sensor nodes (LPNs) have non-linearities and uncertainties due to a poor measuring accuracy \cite{spinelle2015field}. The simpler calibration methods such as polynomial curve fitting methods \cite{badura2018evaluation}, utilized in past could not produce the desired results because of the complexity of the problem. Support vector machines have also been employed in this field \cite{laref2018support}. 
K-Nearest Neighbors and random forest approaches are also being adopted in the past \cite{ferrer2020multisensor}. In \cite{badura2019regression}, the methods of uni-variable and multi-variable neural network regression techniques were implemented. However, those methods did not incorporate the cross-sensitivity between different air quality parameters. In \cite{zaidan2020intelligent}, the authors have implemented some complex non-linear models, however, they are limited to only two pollutants and the methods are less practical and more analytical. 

The past work is unable to address the issues of frequent downtimes and intermittent data-stream at any given time. To overcome the gaps in past work, we aim to present a modular air quality calibration method which readily adjusts according to the availability of measurement sources. Additionally, the design considers the inter-dependency of the pollutants and the climatic parameters. Due to the reliability limitations of low-cost sensors, the data required for incorporating cross-sensitivity may not be available at all times. This demands an adjustable design for the platform that consists of a multi-topic MQTT (Message Queuing Telemetry Transport) broker \cite{atmoko2017iot}. Model-based application of MQTT are explained in \cite{tanabe2020model}. We leverage the real-time monitoring through this protocol similar to the approach pursued in \cite{atmoko2017iot}. The to and fro data transfer helps to get the best-learned weights to the edge module inside the sensor nodes and enables the data to be received at the local database and data processing servers. A calibrator is programmed on each sensor node that comprises of four stages depending upon the data from various sensors. All the stages are powered by trained machine learning agents. The first stage considers each air quality measure separately; the second considers the environmental parameters such as temperature, pressure, and relative humidity. The third stage is designed to incorporate the cross-sensitivity of different air quality factors for the gas-based sensors. Lastly, the fourth layer combines both the gas sensor data and particulate matter data to maximize the information gain because there is evidence of cross-correlation between them \cite{fu2020investigating}. A specialized scheme is adopted to select the models through a micro-controller. We extend our method to provide near-accurate predictions of the future as well. For that reason, we apply a mutual information score-based feature selection and long short term memory-based learning agents.

The rest of this paper is organized as follows. Section II gives details of MAQ-CaF, i.e., our calibration and prediction scheme. Section III entails the modeling and problem definition. Section IV and V delve into the calibration methodology and model selection, respectively. Section VI discusses the forecasting methodology, Section VII provides the validation of the proposed methods, and section VIII concludes the paper.

\section{Calibration and Monitoring}

We propose a decentralized calibration method where the sensor nodes are capable of displaying the pollutant concentrations locally as well as populating sensor values on the remote data integration server. 

IoT technology is the backbone of such a strategy. Bi-directional communication mode is adopted, and its security is studied in detail in \cite{andy2017attack}. In this scheme, one topic of MQTT is reserved for each direction. The publishers are the sensing nodes while the subscribers are the backup and cloud-integration databases, for one of these directions. Two databases enhance the redundancy to avoid data loss in case of any contingency and cyber-attack. Machine learning models are used for learning the calibration models in the cloud. Moreover, integrated cloud also behaves as a publisher, and the subscribers are the sensor nodes. The weights of the learned models are updated using this channel. Once the weights matrix is updated, the calibrated pollutant concentrations are populated on the sensing nodes, immediately. The weights need an update rarely as compared to pollutant concentration data being transmitted to the databases.  

Fig. \ref{iot} provides a detailed illustration of the calibration and forecasting scheme. A multi-topic broker helps to provide services for air quality parameters as well as the calibration models and their weights. For each node, the PM and gas pollutants' concentrations are recorded and the micro-controller loads the trained calibration agent from cloud integration layer onto the electronically erasable programmable read-only memory (EEPROM). The calibrated readings are published to the MQTT-broker.

\begin{figure*}[t]
\centering
\includegraphics[width=\linewidth]{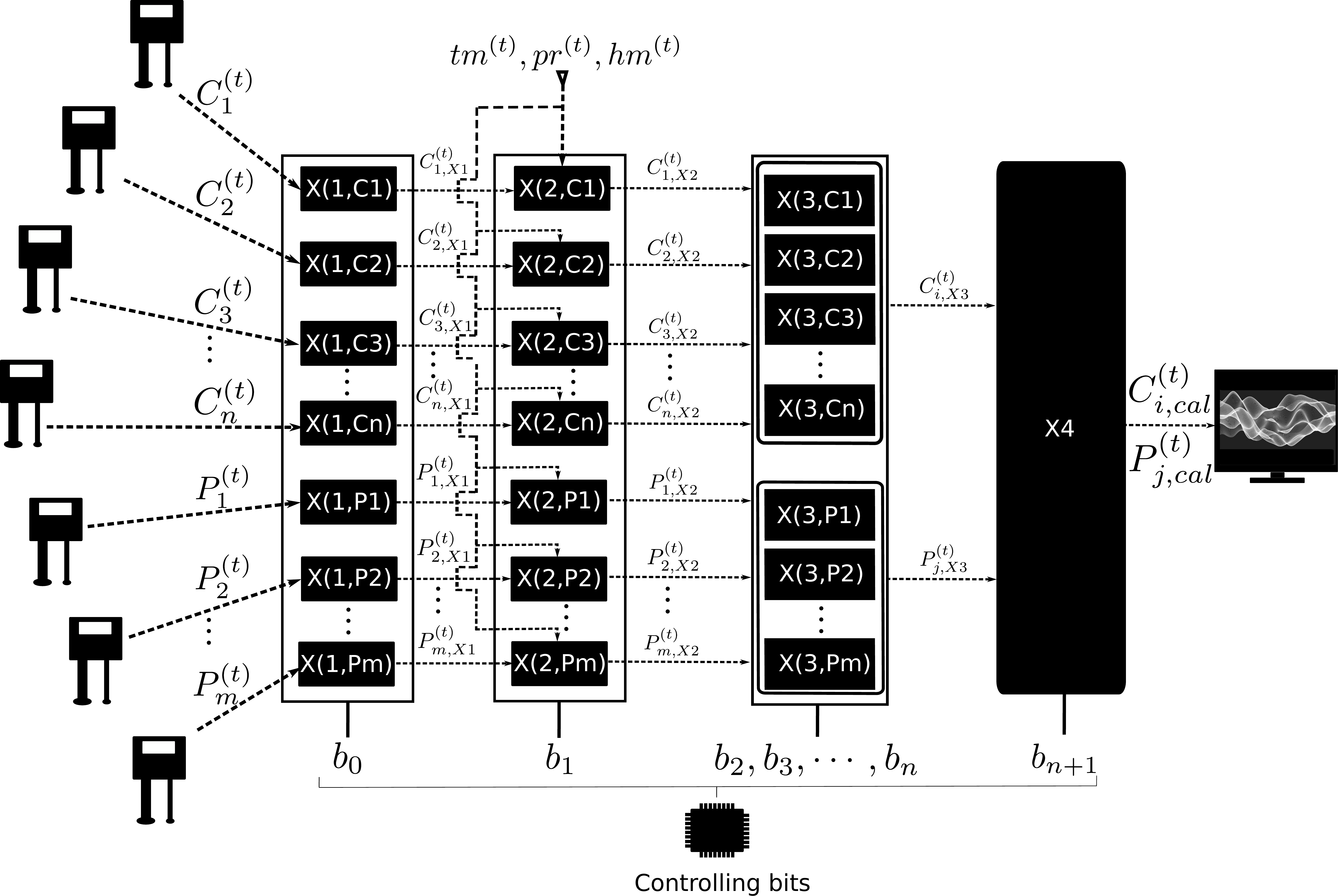}
\caption{Architecture of the proposed MAQ-CaF scheme. Here, the trained agents are labelled as X(stage number, input air quality metric). The modular technique involves multiple calibrating agents and a controller. Stage X$1$ requires only the $S_{in}$ to give the estimating of each air quality measure. Stage X$2$ takes in the estimation from the previous stage and environmental factors $tm^{(t)}$, $pr^{(t)}$ and $hm^{(t)}$. Stage X$3$ and stage X$4$ uses multiple pollutants for leveraging cross-sensitivity among them. Any stage can be enabled at any time using its enable controlling bit, as per requirement and availability of data.\vspace{-4mm}}
\label{calibration_scheme}
\end{figure*}

\section{Modelling for Calibration}
The article aims to propose an effective technique for the calibration of LPNs by exploiting the high fidelity data obtained from high-cost (by order of magnitude) and high precision sensor nodes HPNs. With such an objective in perspective, we propose a modular air quality calibrator and forecaster scheme, MAQ-CaF, as illustrated in Fig. $\ref{calibration_scheme}$.  We define the input form a low-cost sensor array comprising of the sensor output from either the gas-based sensors $C_i \in \mathbb{R}^t$ for $i\in[1,2,3,\cdots, n]$ number of sensors, or the particulate matter based sensors $P_j\in\mathbb{R}^t$ for $j\in[1,2,3,\cdots, m]$ number of sensors, where $t\in[1,2,3,\cdots, T]$ indicates all the time points. Similarly, we specify the outputs of the learning mechanism  $C_{i,cal} \in \mathbb{R}^t$ and $P_{j,cal} \in \mathbb{R}^t$. Subsequently, for the convenience we combine the sensor inputs and output as $S_{in}\in[C_i^{(t)},P_j^{(t)}]$ and $S_{out}\in[C_{i,cal}^{(t)},P_{j,cal}^{(t)}]$. To keep the approach modular, the calibration is carried out in four stages $X\in[\textrm{X}1, \textrm{X}2, \textrm{X}3, \textrm{X4}]$, with each stage comprising multiple calibration agents. The agents are learned during a training phase using the historical target data comprising of $C^{*}_i$ and $P^{*}_i$ readings provided by the HPNs. Moreover, weather parameters such as temperature $tm_i$ and $tm_j \in \mathbb{R}^t$, pressure $pr_i$ and $pr_j \in \mathbb{R}^t$ and relative humidity $hm_i$ and $hm_j \in \mathbb{R}^t$ in the surroundings of $i$-th or $j$-th sensor are also leveraged to boost up the calibration performance. The calibrated measurements $S_{out}$ are subject to the forecasting architecture to generate $C_{i,pred}^{(t)}$ and $P_{j,pred}^{(t)}$ for a pre-specified number of days.
\vspace{-4mm}
\subsection{Problem Definition}
\textbf{Problem}: 
design an effective calibration mechanism without compromising on the flexibility of the technique. A generalized approach is presented. 

\textbf{Given}:
\begin{itemize}
\item an LPN $L$ for measuring $n+m$ different pollutants, giving $C_i$ and $P_j$ real-time measurements with frequent downtimes
\item temperature $tm_i$ and $tm_j$, atmospheric pressure $pr_i$ and $pr_j$ and relative humidity $hm_i$ and $hm_j$ measurements with random downtimes, and
\item an HPN $H$ with historical measurements $C^{*}_i$ and $P^{*}_j$ of $n+m$ number of different pollutants. 
\end{itemize}

\textbf{To Find}: 
\begin{itemize}
\item calibrated LPN measurements $C_{i,cal}^{(t)}$ and $P_{j,cal}^{(t)}$ for gaseous and particular matter sensors,
\item modular design of the calibrator to achieve flexibility, and
\item forecast estimates $C_{i,pred}^{(t)}$ and $P_{j,pred}^{(t)}$ for a specified number of days ahead.
\end{itemize}

\section{Calibration Techniques}
MAQ-Caf consists of multiple calibration stages $(\textrm{X}1, \textrm{X2}, \textrm{X3}$, and $\textrm{X4})$ each comprising one or more calibration agents. Once the trained weights of a learned calibration agent are available, the calibration can be applied in real-time depending upon the controllers' output $B\in[0,1]^{n+m+5}$. For training purposes, an LPN node is co-located with an HPN for an extended time period ($\sim$ 6 Months), and the acquired data is logged on the server. This is done for the atmospheric data and the smog constituents: CO, $\textrm{SO}_2$, $\textrm{O}_3$, $\textrm{NO}$ and $\textrm{PM}_x$ concentrations. The time points are averaged over a period of 1 hr for the training purpose. In the following, we provide a brief of each of the calibration techniques evaluated for the proposed MAQ-CaF scheme. 
\vspace{-4mm}
\subsection{Support Vector Regression}
For each sensor $i$ or $j$, there exists a sensor output from both LPN and HPN to be used for training. Each one of them represents a feature in a higher dimensional space generated by the SVR (Support Vector Regression) model, as purported in \cite{wang2019calibration}. A mapping rule  $\mathcal{F}_C$ and $\mathcal{F}_P$ for the measurement of gases and particulate matter is then established between LPN and HPN readings. 
\begin{equation}
    \mathcal{F}_C: C_{i}\rightarrow C_{i,X};
    \mathcal{F}_P: P_{j}\rightarrow P_{j,X}.
\end{equation}
In an SVR model the weights vector $\omega_i$ is indicative of the slope of dividing hyperplane and the feature space is selected to keep the separation at maximum for all time points $t\in[0,1,2,...,T]$. The optimization function is,
\begin{equation}
    \min ~(\frac{1}{2}\norm{\omega_{i}}_2+\gamma\sum_{t=1}^{T}\zeta_{i}^{(t)}),
\end{equation}
where $\zeta_{i}^{(t)}$ controls the hard or soft margin in case of the non-separable environment in the newly created feature space, and $\gamma \in \mathbb{R}^2$ in the range of $[0,1]$. 
The gaussian kernel was selected for the experimentation for CO, $\textrm{SO}_2$, $\textrm{O}_3$ $\textrm{NO}_2$, and $\textrm{PM}_x$. 
\vspace{-4mm}
\subsection{Random Forest}
Random Forest is an ensemble-based algorithm for regression and has been a prudent selection in the past\cite{wang2019calibration} for calibration of $\textrm{PM}_{2.5}$ values. Considering $Q$ number of decision trees, the output is obtained through,
\begin{equation}
    C_{i,X}^{(t)}=\frac{1}{Q}\sum_{q=1}^Q tree^{(t)}_{X}
\end{equation}
A similar procedure repeats for $P_{j,X}^{(t)}$.
We employ standard grid search for evaluating the hyper-parameters through cross-validation.
\vspace{-4mm}
\subsection{K-Nearest Neighbors}
In K-Nearest Neighbor's feature space, the closest $K$ points to $S_{in}$ are used for training purposes, for every $X$. Their Euclidean distances are used to ascertain the $K$ distinct neighboring samples in $\mathbb{K}$. Finally, their mean is evaluated to get the result for both $C_{i,X}^{(t)}$ and $P_{j,X}^{(t)}$. Hence, $C_{i,X}^{(t)}=\frac{1}{k}\sum_{C_{i}^{(t)}\in \mathbb{K}}C_{i}^{(t)}$.

\subsection{Multivariate Linear Regression}
Multivariate Linear Regression is a simple yet promising linear regression technique for multiple feature variables. The weight arrays $\alpha_i$ and $\alpha_j$ for each kind of air quality metric are learnt by minimizing the euclidean distance form the regressed hyper-plane so, $C_{i,X}^{(t)} =\sum_{i}\alpha_i \cdot C_{i}^{(t)}$. The data from gas-based and particular matter-based sensors are subjected to the same procedure.

\subsection{Deep Neural Networks}
Apart from the classical Machine Learning based methods, as discussed above, we also explored deep neural networks. Owing to the richness of information provided by the correlating gases and the particulate matter data, a deep neural network may better recognize the complicated patterns and improve the calibration accuracy.

\begin{equation}
    C_{i,X}^{(t)}=\sum_{i}\phi_{xp} (\Omega_i\cdot C_{i}^{(t)});
    P_{j,X}^{(t)}=\sum_{i}\phi_{xp} (\Omega_j\cdot P_{j}^{(t)})
\end{equation}
where $\phi_{xp}$ represents the activation functions, and $xp$ is the number of perceptrons. Moreover, $\Omega_i$ and $\Omega_j$ indicate the weights learned during the training process. Deep neural networks present an apt solution to such regression problems where there exist highly complex and non-linear patterns in different air quality measures. The activation function types, loss function types, and the number of hidden layers and their perceptron counts are determined through a k-fold cross-validation method.

\vspace{-4mm}
\section{Model Selection}
MAQ-CaF has a modular design comprising multiple stages and their associated calibration agents, as shown in Fig. $\ref{calibration_scheme}$. These calibrating agents are selected from the various learned models described in the last section. Note that these learning agents have different performances at different stages. 
Some agents may show superior performance under fewer input features, while others may be better with denser and larger input feature space. With that in consideration, we propose a model selection approach guided by an evaluation metric to achieve the sub-optimal calibration while preserving the flexibility. 

Having the calibration results for each of the learning method invokes a need for a metric of evaluation. The evaluation of the calibration of environmental sensors has long been performed via the root mean squared error (RMSE) method. If the real measurements from an HPN are $C_{i,X_{real}}^{(t)}$ and $P_{j,X_{real}}^{(t)}$, the error functions are defined as,

\begin{equation}
    \textrm{RMSE}_{C_i}^X = \sqrt{\frac{1}{T}\sum_{t=1}^{T}(C_{i,X}^{(t)}-C_{i,X_{real}}^{(t)})^{2}},
\end{equation}
\begin{equation}
    \textrm{RMSE}_{P_j}^X = \sqrt{\frac{1}{T}\sum_{t=1}^{T}(P_{j,X}^{(t)}-P_{j,X_{real}}^{(t)})^2}.
\end{equation}

In stages X$1$ and X$2$, all agents are trained using each of the proposed calibration method, individually. We adopt a specially designed optimization function for selecting the model with the least $\textrm{RMSE}_C$ and $\textrm{RMSE}_P$, accordingly. For all the trained models $m$, $M_{i}\in\mathbb{R}^m$ has the RMSE values of those models for $C_i$ inputs. Similarly, $N_{j}\in\mathbb{R}^n$ holds the RMSE values of the $n$ trained models for $P_j$ inputs of stage $1$. The following optimization function delivers the argument of the model with the best performance in stage $\textrm{X}1$. 
\begin{equation}
    \argmin_{M_i}~ b_0\cdot\sum_{M_i} (\textrm{RMSE}_{M_i})
    \label{l1}
\end{equation}
\begin{equation}
    \argmin_{N_j}~ b_0\cdot \sum_{N_j}(\textrm{RMSE}_{N_j})
    \label{l2}
\end{equation}
where $b_0$ is the controlling bit which can either be $0$ (stage is disabled) or $1$ (stage is enabled).

For stage X$2$, the process of model selection is repeated in the same way, with additional features being appended namely $tm^{(t)}$, $pr^{(t)}$, and $hm^{(t)}$ for incorporating the environmental factors. That is because the literature claims the interdependence of air quality metrics and environmental factors. Hence, when the bit $b_1$ is set to $1$ the outputs are populated as $C_{i,X2}^{(t)}$ and $P_{j,X2}^{(t)}$.

Stage $3$ is unique in its design and implementation because here, all the data available for all the $n$ gasses is taken in to consideration. Since we are aiming for a modular design, we consider the likelihood of not having the incoming data for all the  gases because of any logistical, cost and network accessibility constraints. In such a case, the past estimation $C_{i,X2}^{(t)}$ or $P_{j,X2}^{(t)}$ serves as the input and the controller enables one of the two blocks. Bits $b_2$ and $b_3$ control the blocks at stage 3 as depicted in Table \ref{Table1}. 

\begin{table}[!hbt]

\caption{Controlling stage $3$ blocks.}
	\label{stage3}
	\centering
 \begin{tabular}{c  c  c c}   

 \hline\hline
 $b_2$ & $b_3$ & $C_{i,X3}$& $P_{j,X3}$\\ [0.5ex] 
 \hline
 0 & 0 & Disable & Disable \\ 
\hline
 0 & 1 & Disable & Enable \\
\hline
 1 & 0 & Enable & Disable \\
 \hline
 1 & 1 & Enable & Enable  \\
 \hline\hline
\end{tabular}
\label{Table1}
\end{table}
In addition to $b_2, b_3$, stage X$3$ further provides a finer level of control using the controlling bits $b_4, b_5, \cdots, b_{n+m+4}$, for providing increased sensor flexibility through modular design. Stage X$3$ comprises two blocks. One of the blocks has a maximum of $n$ number of inputs, while the other block has a maximum of $m$ number of inputs. Since there exists a likelihood of unavailability of any of the input parameters, we design the calibration mechanism to account for all the possible cases. For the block with inputs $C_{i,X2}^{(t)}$, the total number of possible combinations of inputs is $u$, similarly for the block with inputs $P_{j,X2}^{(t)}$, it is $v$. Then, the total number of models is $w$,

\begin{equation}
    u=\sum_{i=1}^{n}\binom ni-n;
    v=\sum_{j=1}^{m}\binom mj-m,
    \label{uv}
\end{equation}
\begin{equation}
    w=u+v=\sum_{i=1}^{n}\binom ni+\sum_{j=1}^{m}\binom mj-(n+m),
    \label{w}
\end{equation}
where (.) indicates the function for evaluating combinations. Hence, X$3$ can have any possible combination of their respective inputs with the exception of single input features because those cases have already been taken care of in X$1$ and X$2$ stages. Similar to the expressions (\ref{l1}) and (\ref{l2}), we use the RMSE values to select the best of all the pre-trained calibration models for making appropriate predictions.  

 \begin{equation}
    \argmin_{M_i}~ b_2\cdot\sum_{M_i} (\textrm{RMSE}_{M_i}) 
\end{equation}
 \begin{equation}
    \argmin_{N_j}~ b_3\cdot\sum_{N_j} (\textrm{RMSE}_{N_j}) 
\end{equation}

The output of this stage will be the nearly calibrated $C_{i,X3}^{(t)}$ and  $P_{j,X3}^{(t)}$. 
For most of the air quality parameters, this stage can serve as the final state. However, considering the recent findings of cross-sensitivity between the gas and PM sensors, there can be another stage, taking advantage of the cross-correlations between $\textrm{PM}_x$ and gaseous pollutants of the LPN. The enable bit $b_{n+m+5}$ helps to bypass ($0$) or activate ($1$) the single block X$4$ of stage X$4$. The process for model selection for this stage is similar to that of stages X$1$ and X$2$, where the calibration models were selected through the RMSE minimization. 
Since only a single block exists at stage X$4$, for each $C_{i,X4}$ a total of only $m$ agents are trained and only one of them is selected for making future inferences, in case of each input to deliver the result of either $C_{i, cal}^{(t)}$ or $P_{j, cal}^{(t)}$.   

This article proposes a methodology that is highly modular and flexible in nature. It can effectively be reduced to any level of learning without a significant rise in the RMSE (drop in performance) as demonstrated later in Section VII. The method is successful in delivering reliable calibration results even under the constraints such as unavailability of other air quality parameters or climatic factors. These characteristics render the proposed solution to be flexible. It also enables us to exploit the simple as well as complex machine learning capabilities which were mostly ignored in such a problem in the past, such as the introduction of deep learning that has proven to be most effective under rich data scenarios. The effectiveness of the methodology is trialled through rigorous testing and validation as demonstrated later in Section VII.
\begin{figure}[t]
\centering
\includegraphics[width=\linewidth]{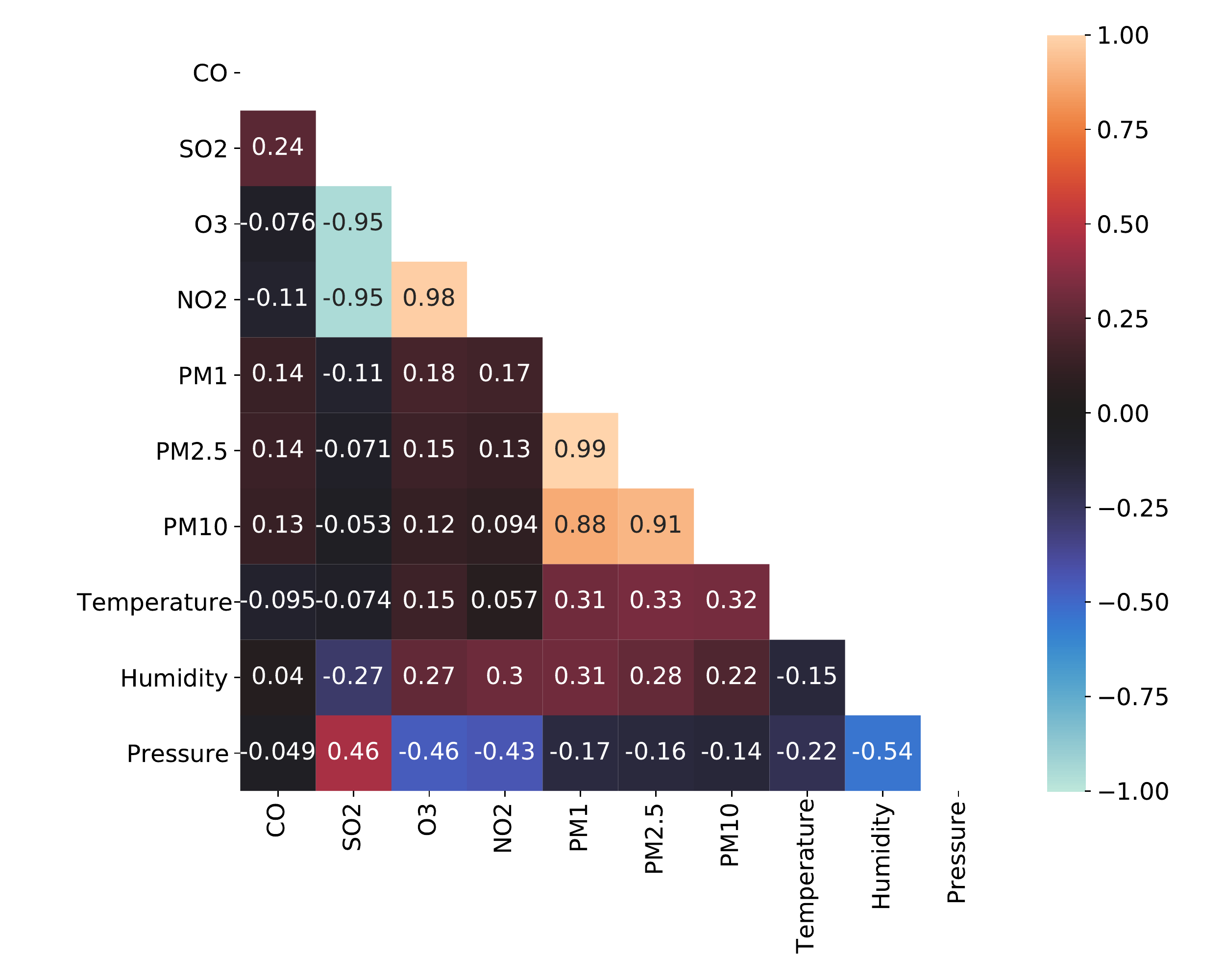}
\caption{Pearson's Correlation Coefficients corresponding to different air quality parameters. }
\label{corr} 
\end{figure}

\begin{figure}[t]
\centering
\includegraphics[width = \linewidth]{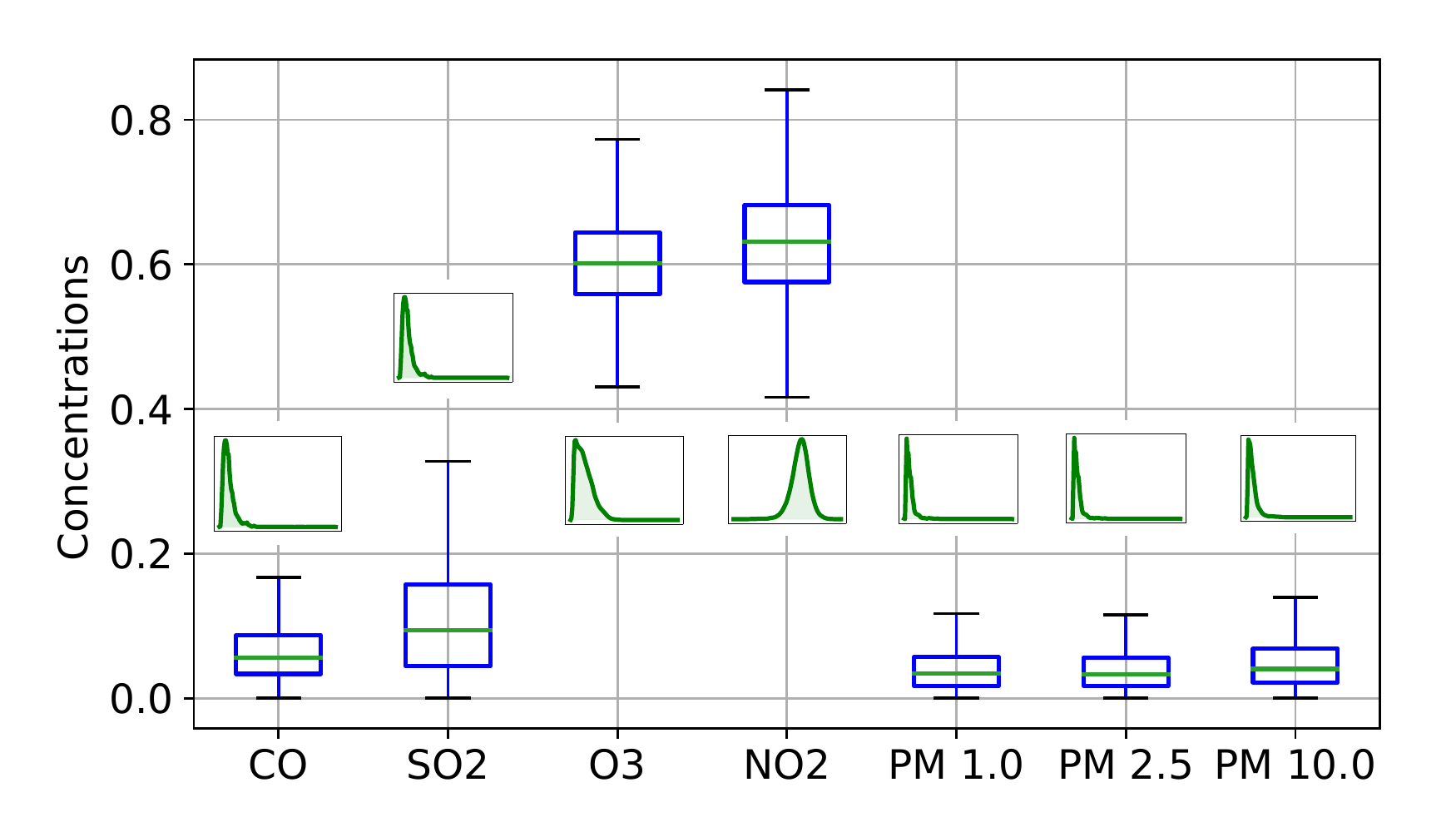}

\caption{Distribution and box-whisker plots of the normalized air quality metrics' concentrations for comparative analysis. \vspace{-4mm}}
\label{boxer} 
\end{figure}

\section{Forecasting Through LSTM}
The calibrated measurements $C_{i, cal}^{(t)}$ and $P_{j, cal}^{(t)}$ are stored in the cloud integration database where the data is further processed to predict the future trends of the air quality parameters. Such an analysis is instrumental in taking preemptive measures, accordingly. LSTM-based networks are widely employed for time series predictions. We propose to establish a generic architecture for determining effective feature set. The correlation matrix provides an insight into the linear cross-correlations only, whereas, $C_{i, cal}^{(t)}$ and $P_{j, cal}^{(t)}$ of different air quality parameters have non-linear correspondence as well. To capture the non-linear dependencies, we propose a solution based upon the mutual information score $I_{(s,s')}$ between any two features in the feature pool, as in \cite{Maciejewska2014IndoorAQ}. Since $I$ has the capability to capture the non-linear dependencies also, it is a naturally preferred choice. $H(.)$ gives the entropy of a random variable,
\begin{equation}
    I_{(s,s')} = H(S_{s})+H(S_{s'|s'\neq s})-H(S_s,S_{s'|s'\neq s}), 
\end{equation}
where $s\in[1,2,\cdots,n+m]$ is the subscript of the air quality parameter to be forecast and $s'\in[1,2,\cdots,n+m+3]$ is the representative of all the other features in the feature pool to include all the gaseous and particulate matter air quality parameters along with the three environmental factors of temperature, pressure and relative humidity. Our goal is to select the features having highest mutual information scores with the parameter to be forecast. Therefore, for each air quality parameter $s$,

\begin{equation}
    \argmax_{|L|=l} \sum_{s'\in L}I_{(s,s')},
\end{equation}
we get the $l$ features with highest mutual information scores with the air quality parameter to be calibrated. The selected number of features $l$ depends upon the computational capabilities and resource availability. It is advised to adopt a cross-validation mechanism to ascertain that number.

\begin{figure}[ht]
\centering
\includegraphics[width = \linewidth]{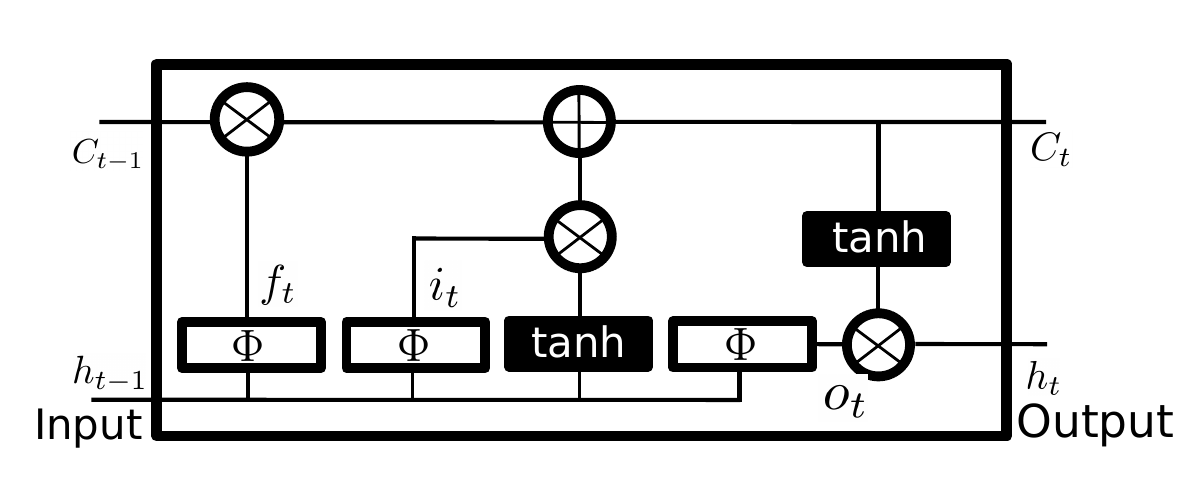}

\caption{A single unit of LSTM carrying tanh and sigmoid activation functions.}
\label{lstm} 
\end{figure}

We get the features and feed them as input to the multivariate-LSTM for making predictions. In \cite{article1234}, a comprehensive methodology is proposed focused on modelling LSTM for forecasting of air quality measures. We obtain the output $h_t$ of the LSTM through its cell state, that combines the cell state $C_t$ and output gate $o_t$, as depicted in Fig. \ref{lstm}. We denote weights matrix by $W_C$ and biases by $b_C$. 

\begin{equation}
    C_t=f_t\cdot C_{t-1}+i_t \cdot tanh(W_{C}[h_{t-1},X_t]+b_C),
\end{equation}

\begin{equation}
    h_t=o_t \cdot tanh(C_t)
\end{equation}
Finally, fully connected layers are added to the network to train the agent to get $S_{pred}^{(t)}\in[C_{i,pred}^{(t)},P{j,pred}^{(t)}]$.

\begin{figure}[t]
\centering
\includegraphics[width = 2.5in]{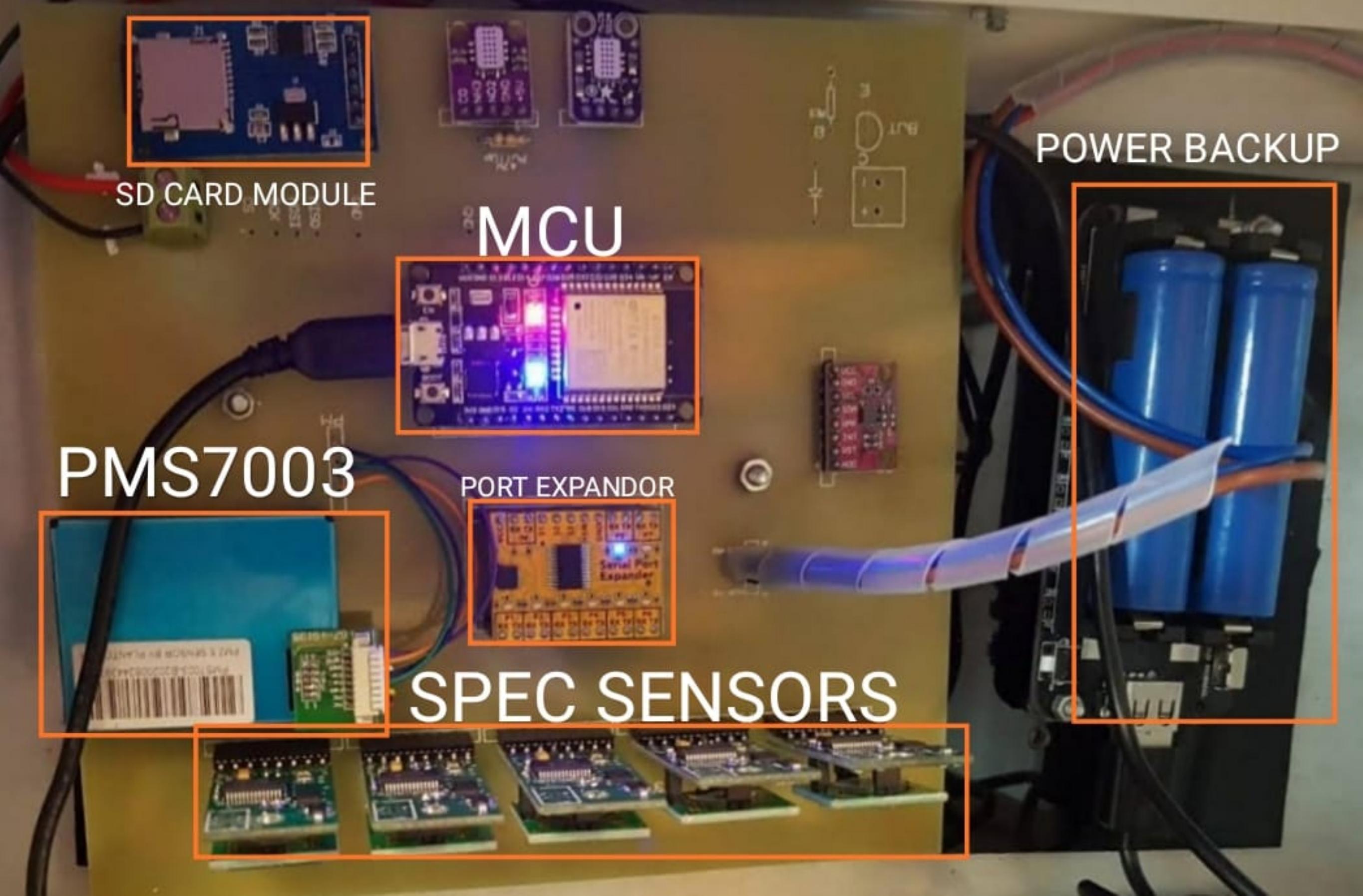}

\caption{A low-cost sensor node containing SPEC and PMS $7003$ sensors along with the communication assembly (ESP32) capable to interacting with MQTT-broker. It has a micro-control unit (MCU), local storage capability in the form of SD-card module, and a power backup. \vspace{-4mm}}
\label{node} 
\end{figure}
\vspace{-6mm}
\section{Numerical Validation}
With a well-defined model of the proposed calibrator and forecaster, we proceed to validate the efficacy and reliability of the methods as discussed in the previous sections for co-location-based calibration and forecasting.

\begin{figure*}[ht]
\centering
    \centering
  \subfloat[\label{6a}]{%
       \includegraphics[width=0.3\linewidth]{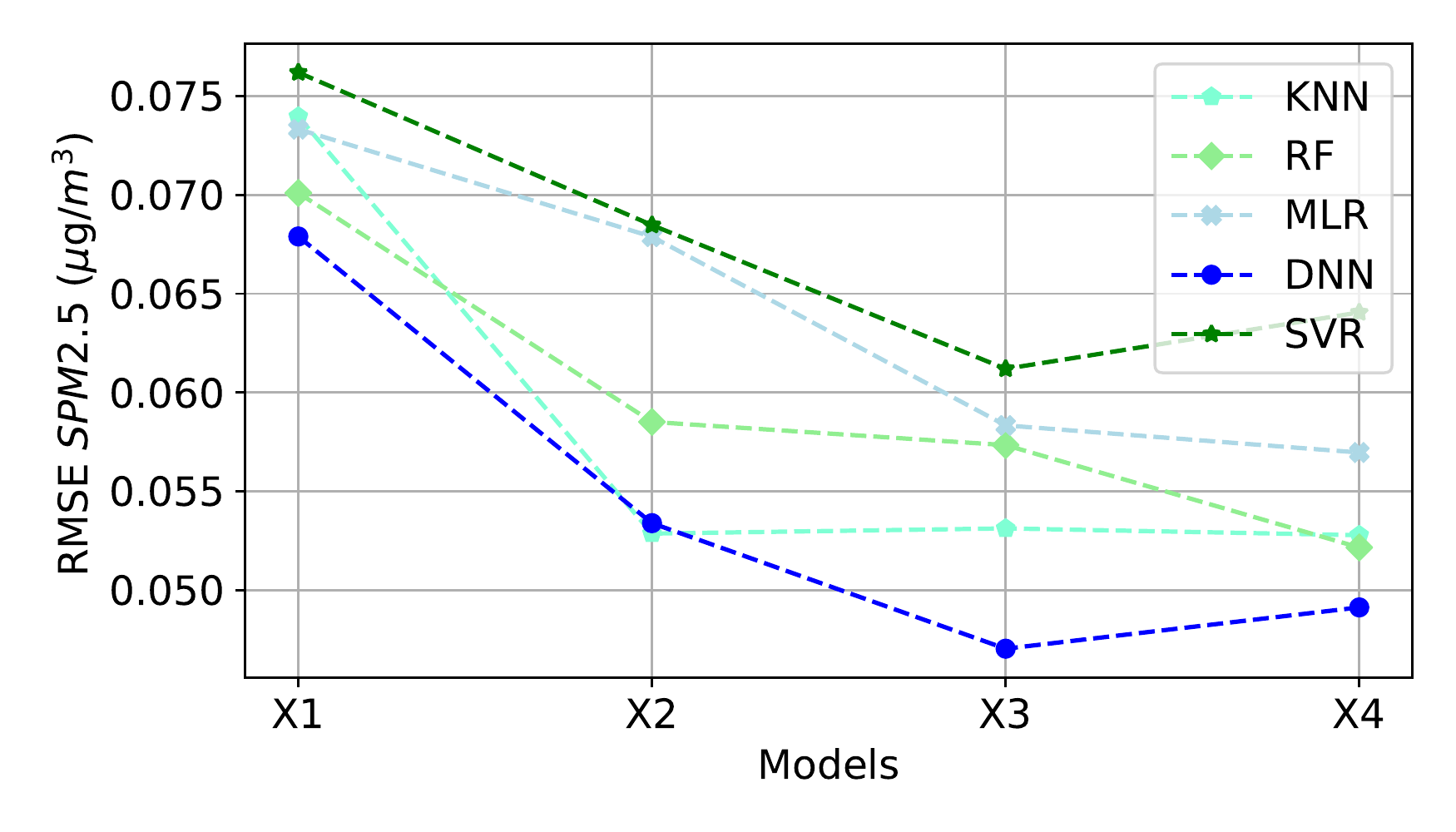}}
       \hfill
  \subfloat[\label{6b}]{%
        \includegraphics[width=0.3\linewidth]{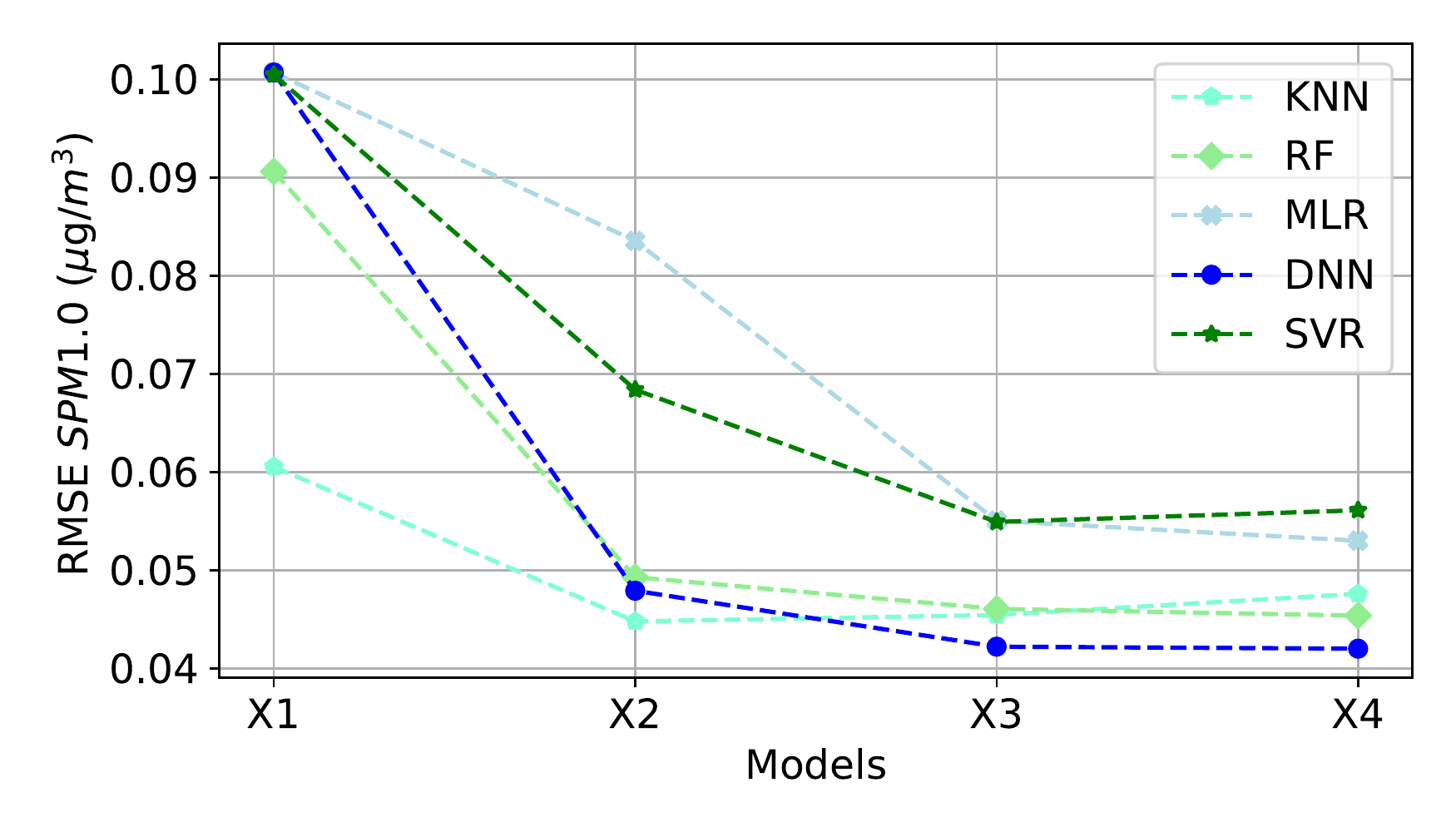}}
          \hfill
          \subfloat[\label{6c}]{%
        \includegraphics[width=0.3\linewidth]{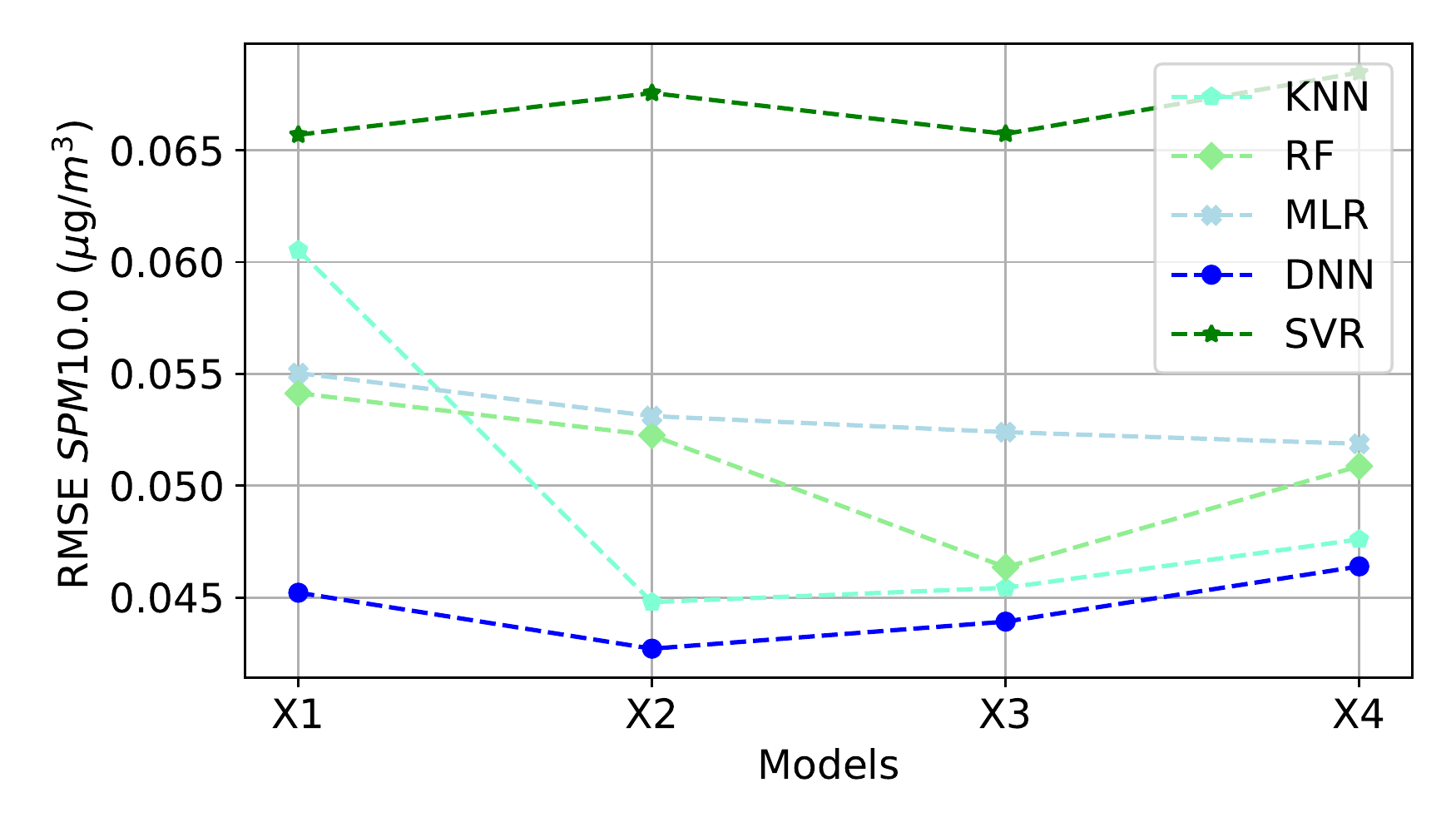}}
                 \hfill
                \subfloat[\label{6d}]{%
        \includegraphics[width=0.3\linewidth]{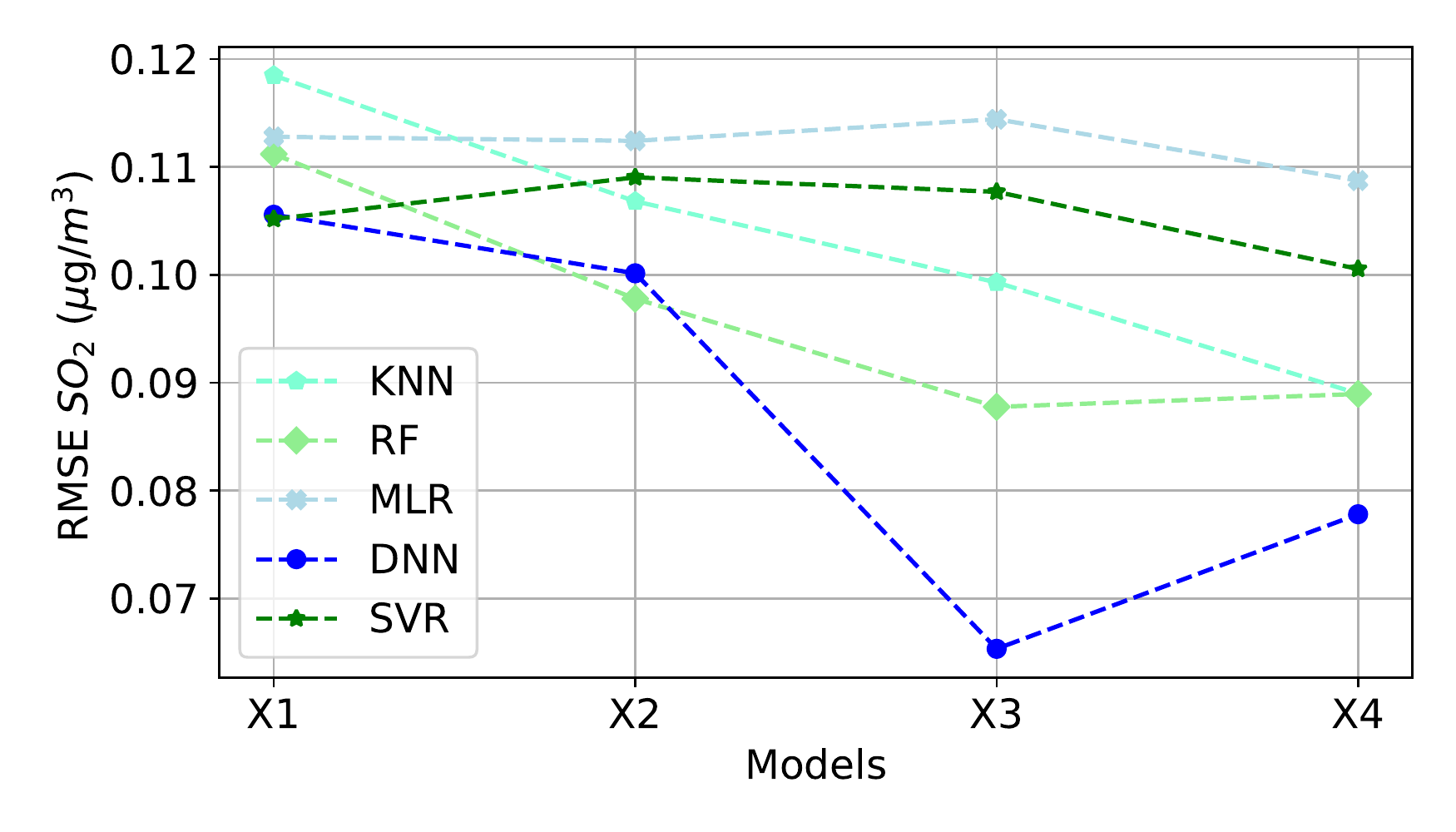}}
        \hfill
        \subfloat[\label{6e}]{%
        \includegraphics[width=0.3\linewidth]{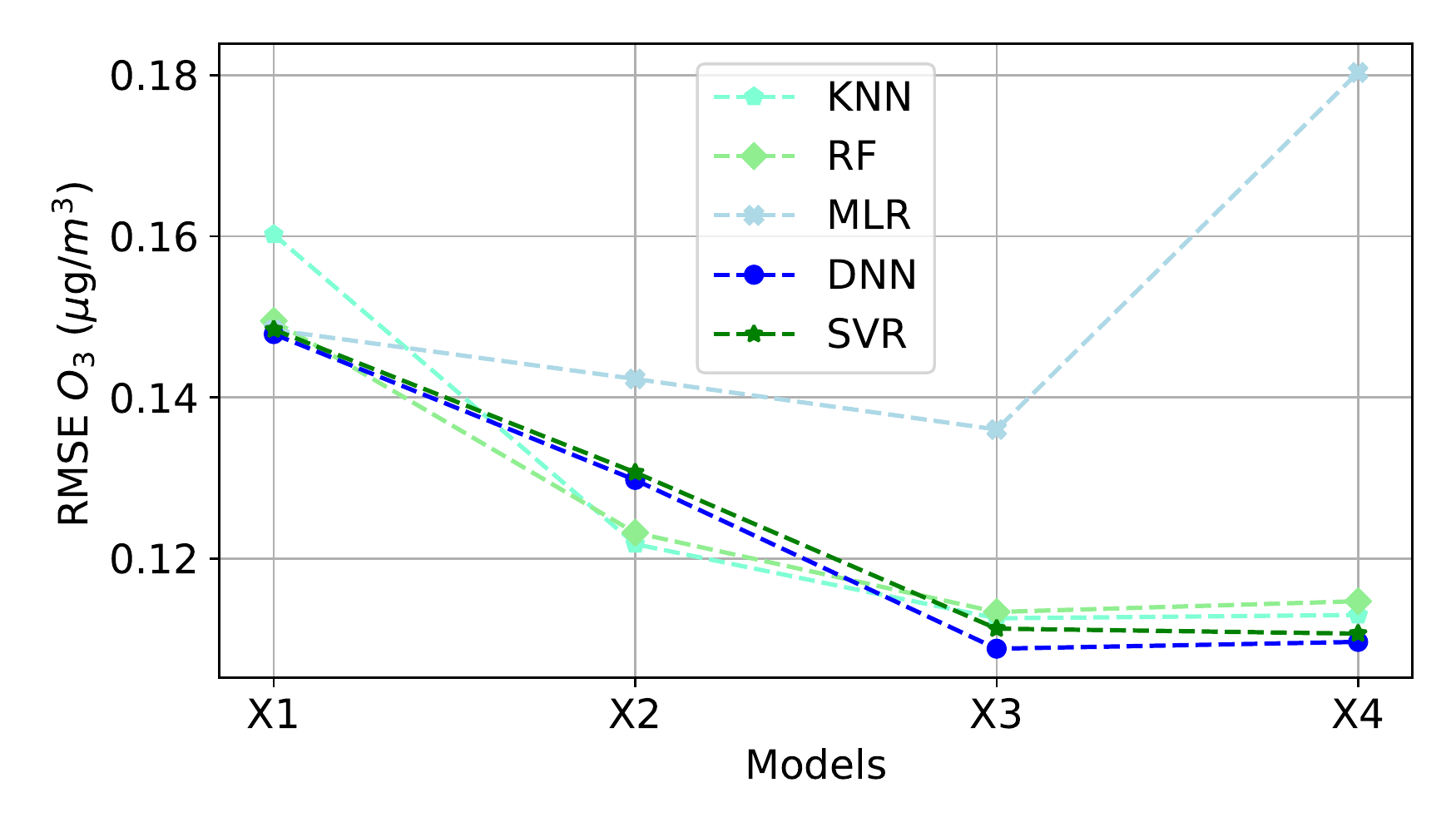}}
                 \hfill
                     \subfloat[\label{6f}]{%
        \includegraphics[width=0.3\linewidth]{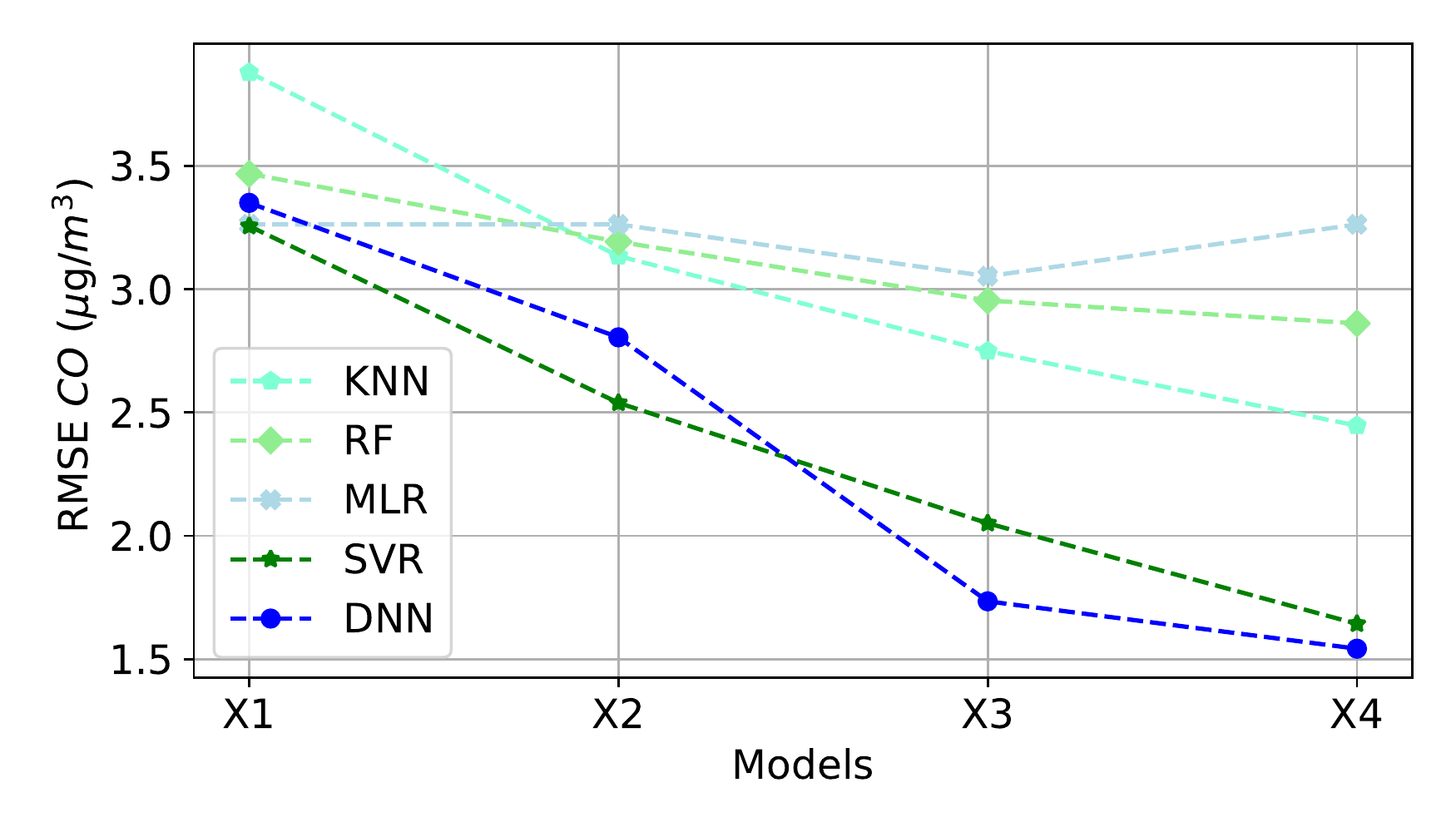}}
                 \hfill
                                      \subfloat[\label{6g}]{%
        \includegraphics[width=0.3\linewidth]{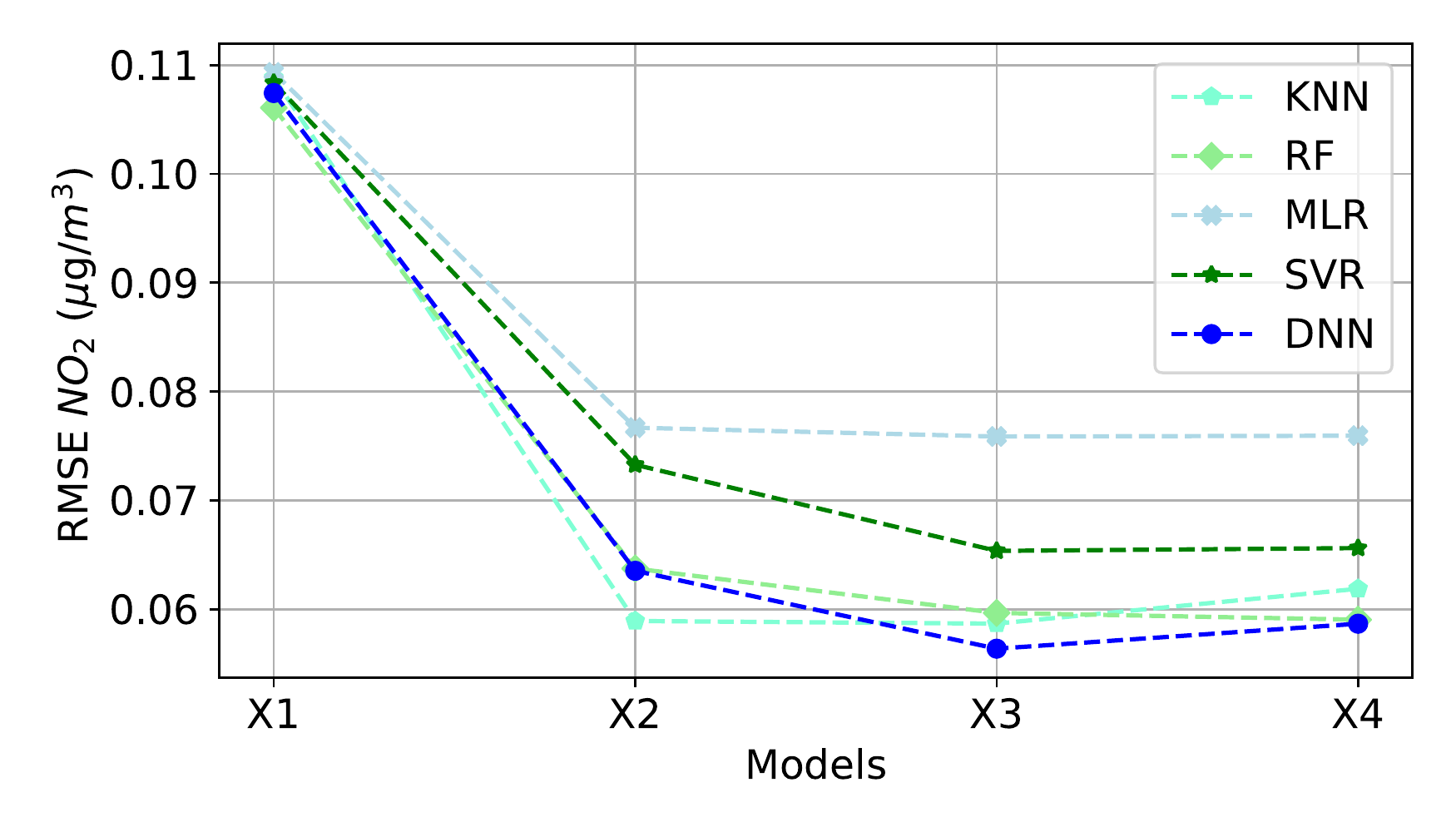}}
                 
 \caption{RMSE values of all the models at levels X1, X2, X3, X4 corresponding to the respecting air quality parameter.\vspace{-4mm}}

  \label{rmse} 
\end{figure*}

\begin{figure*}[ht]
\centering
    \centering
  \subfloat[\label{7a}]{%
       \includegraphics[width=0.3\linewidth]{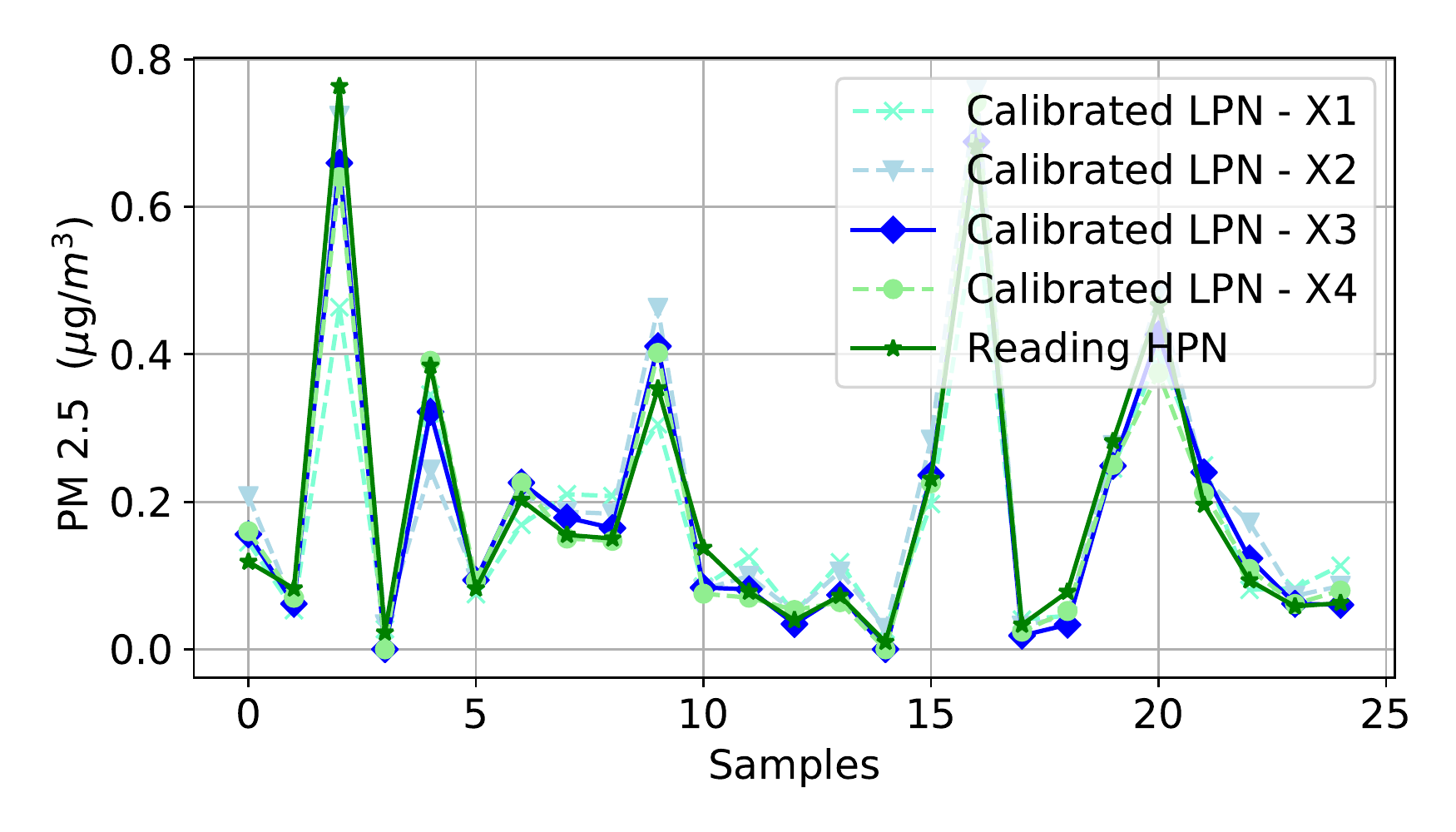}}
       \hfill
  \subfloat[\label{7b}]{%
        \includegraphics[width=0.3\linewidth]{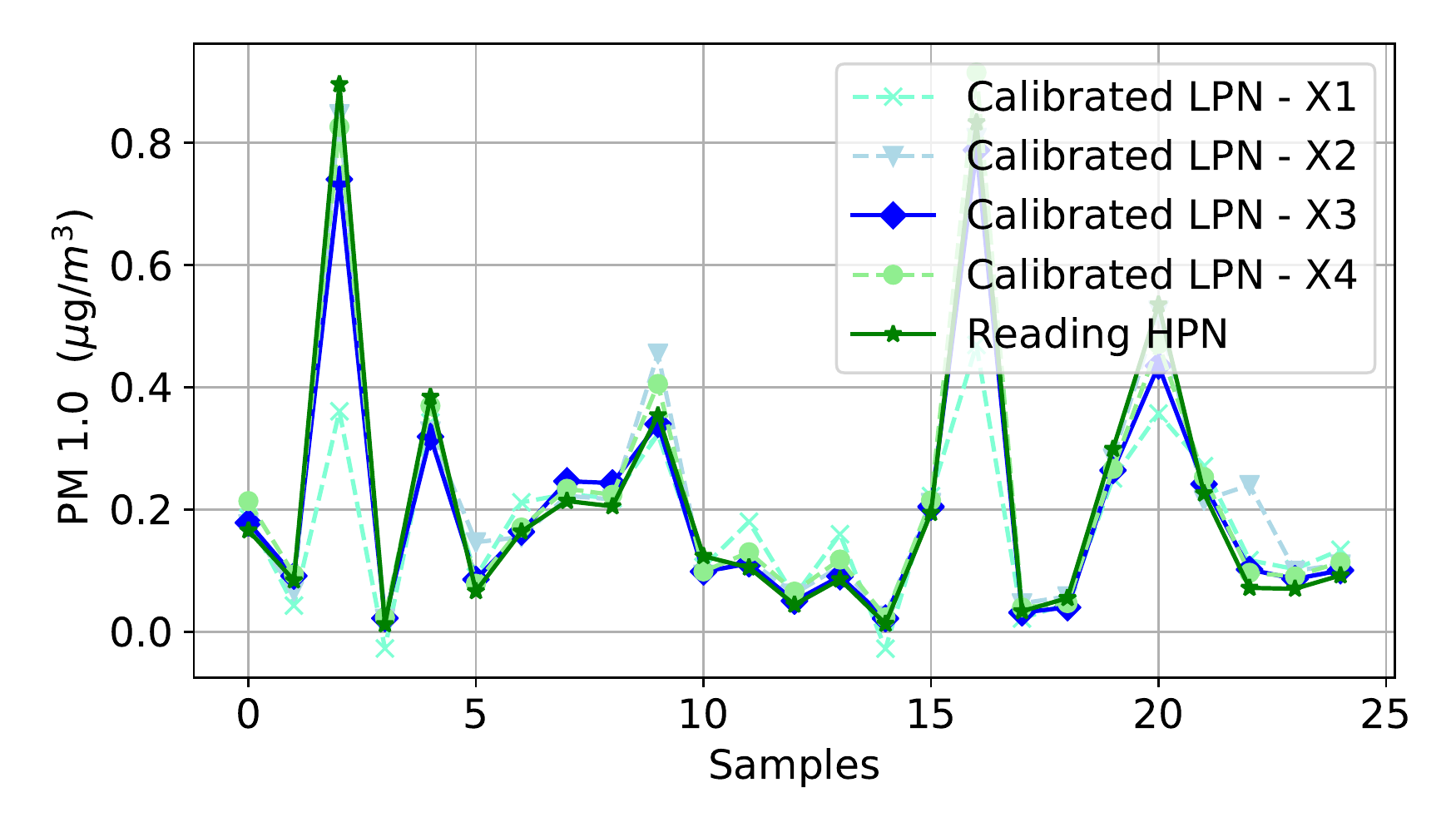}}
          \hfill
          \subfloat[\label{7c}]{%
        \includegraphics[width=0.3\linewidth]{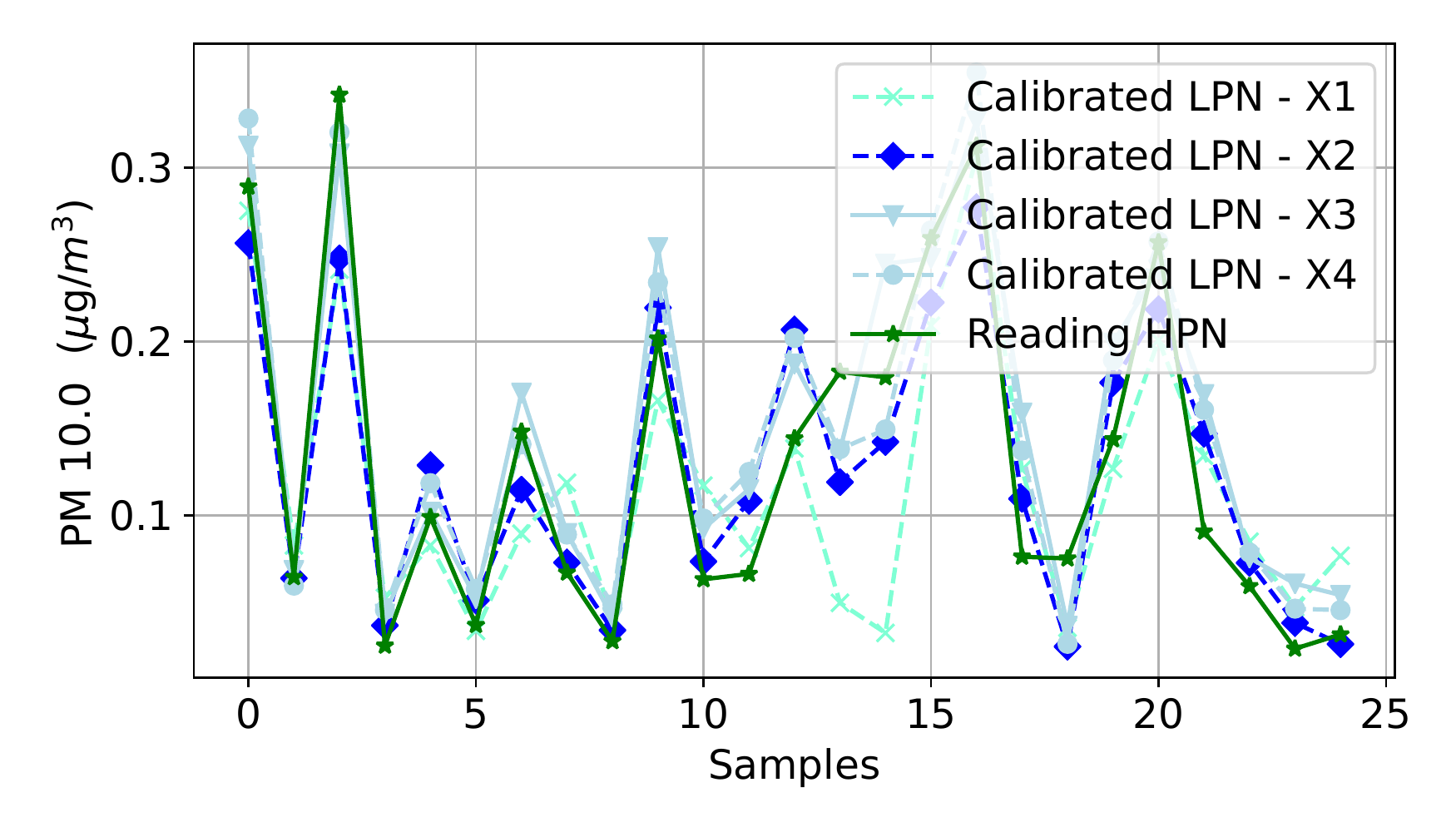}}
                 \hfill
                \subfloat[\label{7d}]{%
        \includegraphics[width=0.3\linewidth]{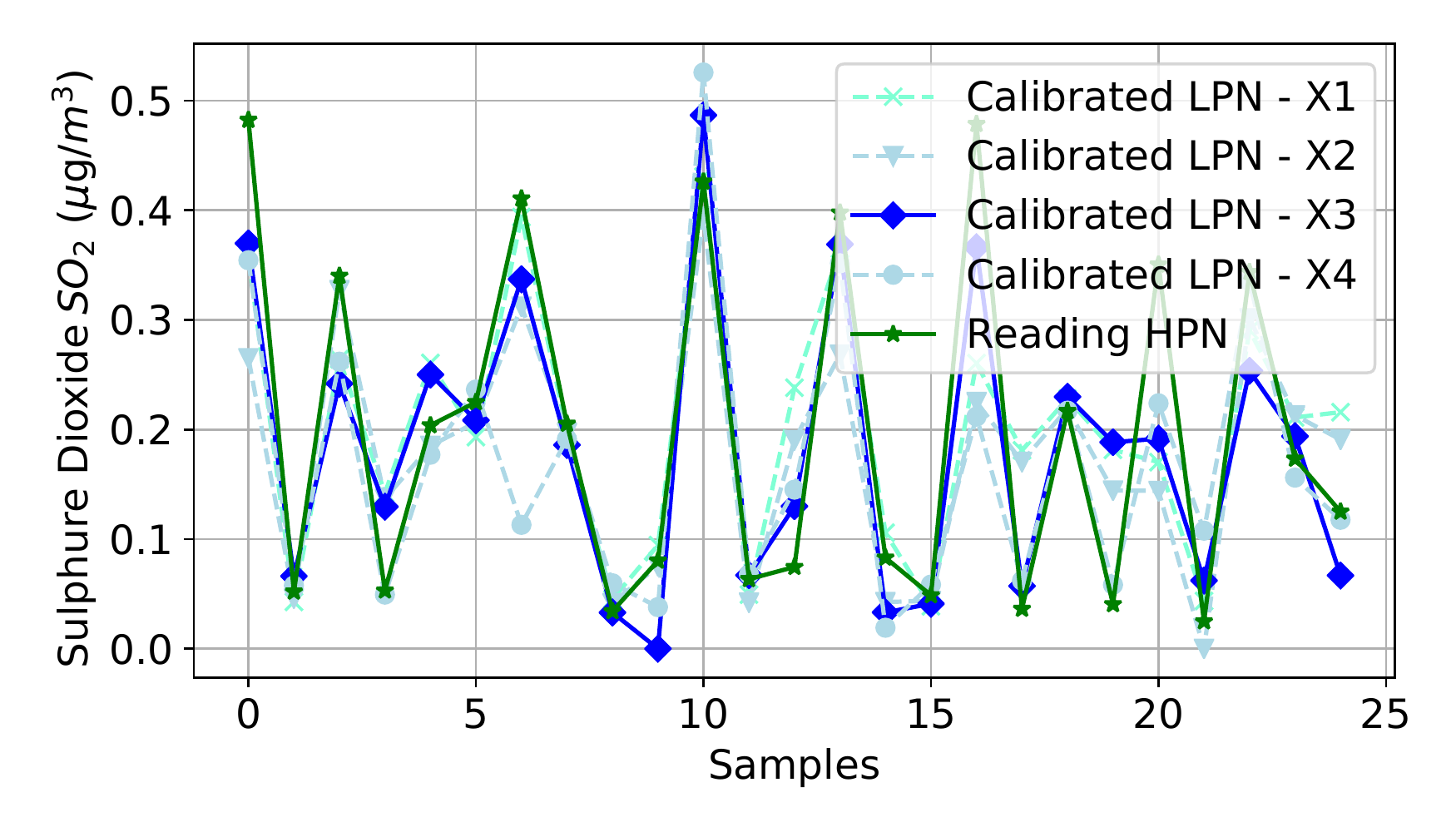}}
        \hfill
        \subfloat[\label{7e}]{%
        \includegraphics[width=0.3\linewidth]{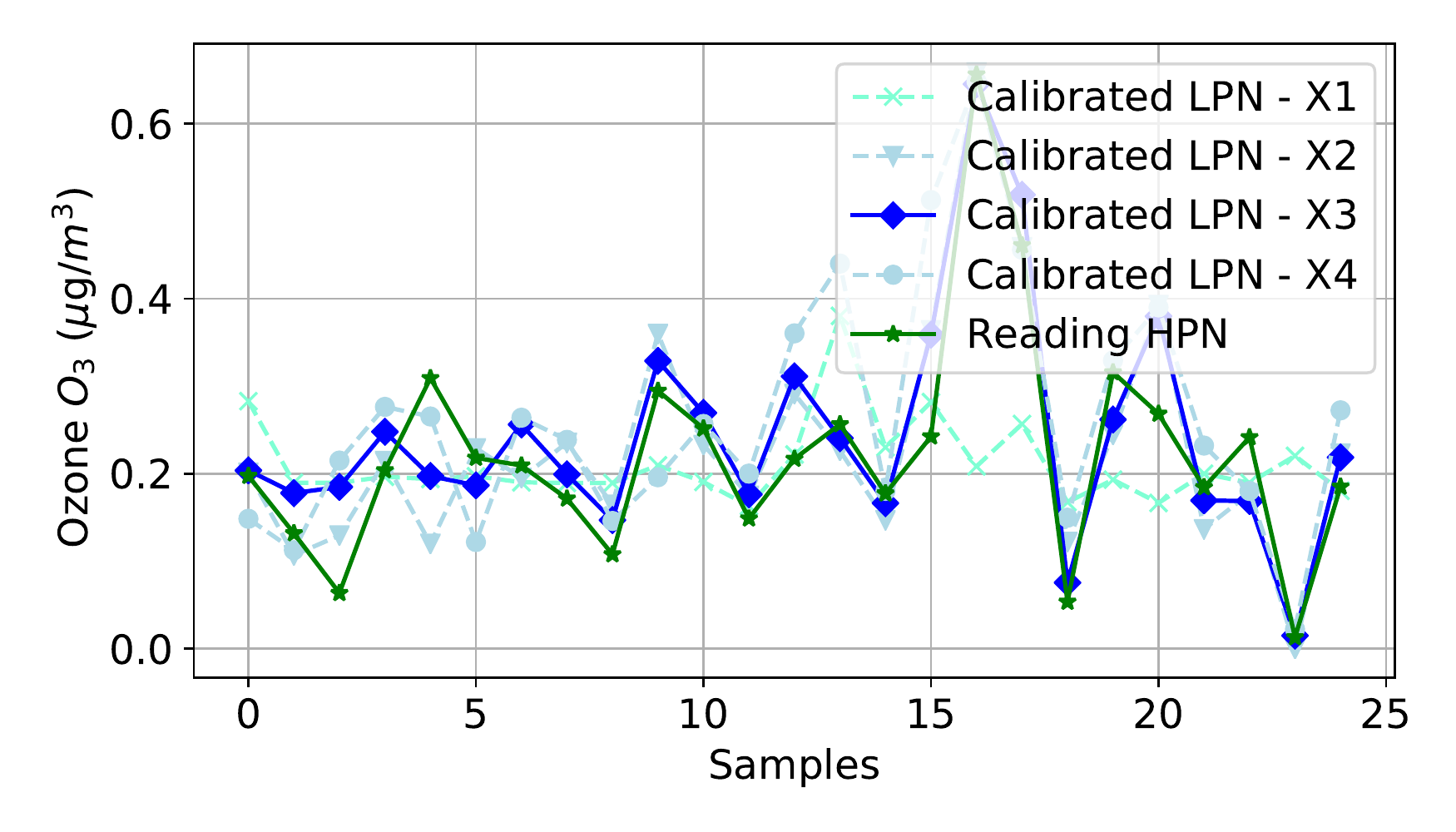}}
                 \hfill
                     \subfloat[\label{7f}]{%
        \includegraphics[width=0.3\linewidth]{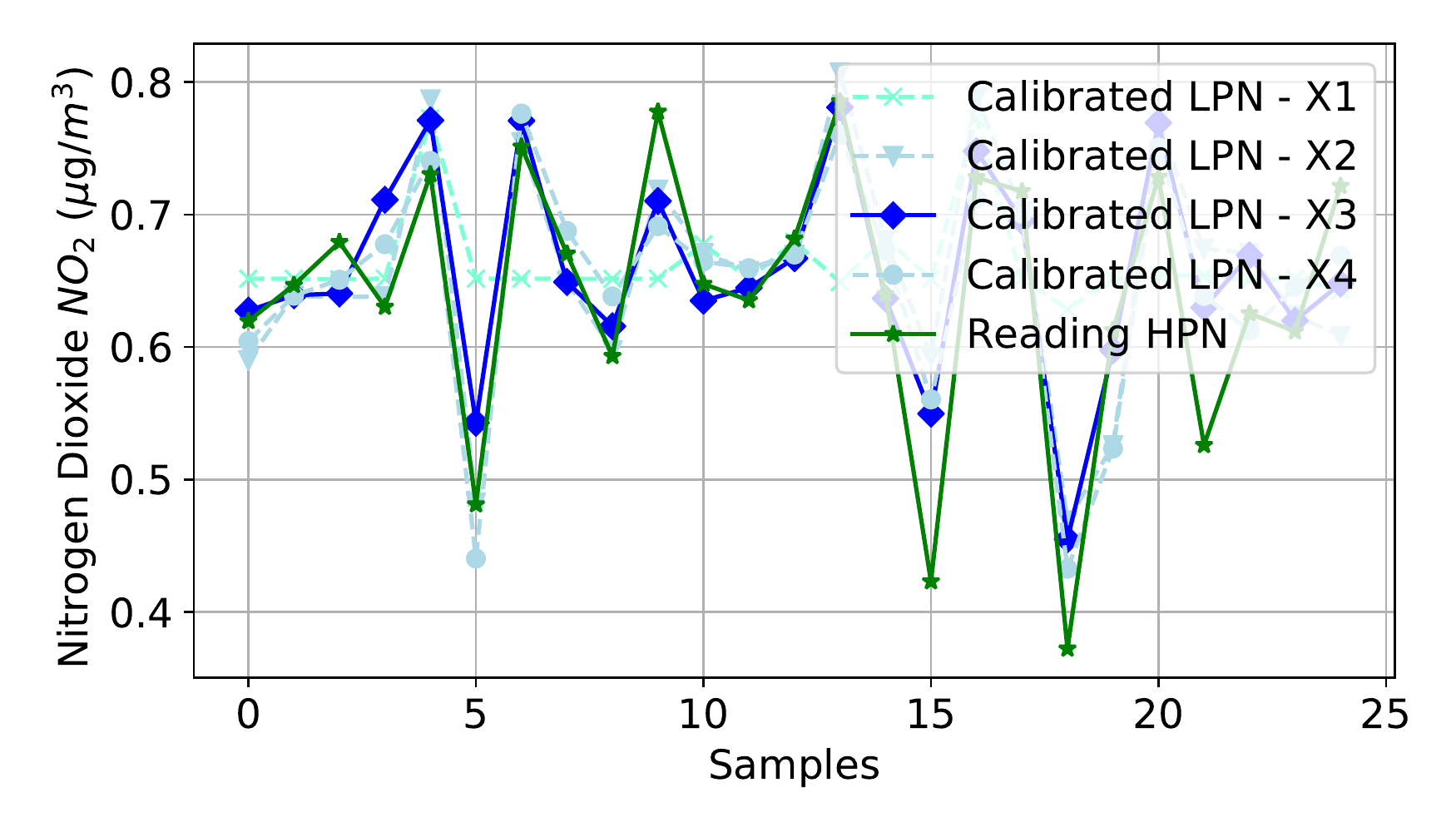}}
                 \hfill
                                      \subfloat[\label{7g}]{%
        \includegraphics[width=0.3\linewidth]{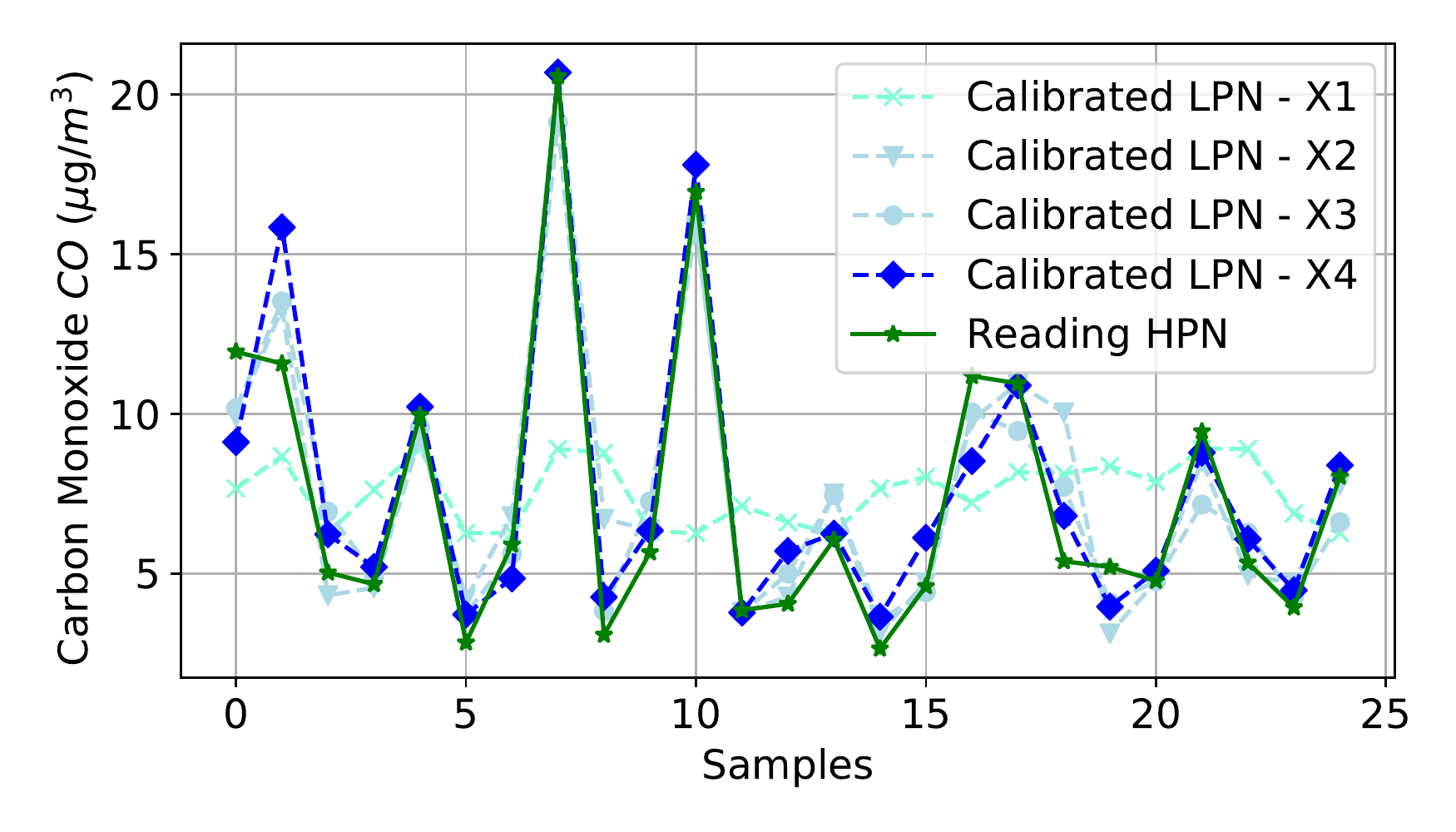}}
                 
 \caption{Time samples of calibrated signals at stages X$1$, X$2$, X$3$, X$4$ corresponding to the respecting air quality parameter.}

  \label{samples}
  \vspace{-4mm}
\end{figure*}
\vspace{-4mm}
\subsection{Developing Sensing System}
We design a low-cost sensing system by leveraging digital electro-chemical SPEC Sensors (DGS-CO-$968-034$, DGS-NO2-$968-043$, DGS-SO2-$968-038$, and DGS-O3-$968-042$) . Moreover, we mount the PMS $7003$ on each sensor node assembly. Such sensors determine the particulate matter concentration by optical variations. Their low cost and better spatio-temporal resolution makes them a suitable fit for this study \cite{wang2019evaluating}. To reduce the cost of production, we employ an ESP$32$ micro-controller. We utilize a port extender for Universal Asynchronous Receiver/Transmitter (UART) communication to overcome the port limitation challenge. Fig. \ref{node} shows our developed LPN, a low-cost sensing node with components as referenced in Fig. \ref{iot}. Furthermore, we employ
two co-located Libellium Smart Environment Pro kits \cite{MSU-CSE-06-2} as HPNs. The factors for selecting that product is its ease of IoT integration, longer life, and performance stability guarantees. 
The sensor nodes are located at $31^\circ28^\prime 32.82^{\prime\prime}$N and $74^\circ20^\prime33.13^{\prime\prime}$ in Lahore, Pakistan. The LPN and HPN are incorporated in the same installation. We recorded and used the measurements of $6$ months from February, 2019 to July, 2019 for the analysis and validation. Table \ref{dat} provides the detailed description of the data under consideration. LPN is the low precision node, and HPN-$1$ and HPN-$2$ are the two high precision nodes. 

\begin{table}[!hbt]

\caption{Data Description.}
	\centering
 \begin{tabular}{c | c | c | c}   

 \hline\hline
 Source & Approx. Resolution & Data Dropout & Bad Data\\ [0.5ex] 
 & (measurements/minute) &Approx.$\%$ & Approx. $\%$\\ 
 \hline
 LPN & $0.5$  & $8$& $12$ \\ 
\hline
HPN-$1$ & $1$ & $2$& $<1$ \\
\hline
HPN-$2$ & $1$ & $<1$& $1$ \\
 \hline\hline
\end{tabular}
\label{dat}
\end{table}

The calibration mechanisms as adopted in the past lack the information enrichment capability through the interdependence of the chemical components that constitute the smog. To overcome such issues, we collect the data of past $6 \textrm{ months}$ for different air quality metrics including $\textrm{CO}, \textrm{ SO}_2, \textrm{ NO}_2, \textrm{ O}_3, \textrm{ PM}_{1.0}, \textrm{ PM}_{2.5} \textrm{ and} \textrm{ PM}_{10}$. We conduct a correlation analysis of the air quality metrics to validate the claim using HPN data. Fig. \ref{corr} illustrates the heat map of Pearson's correlation among the considered environmental as well as air quality parameters. The heat map shows that there exists a strong correlation between some of the features, e.g., a high value of $0.95$ exists between $\textrm{NO}_2$ and $\textrm{SO}_2$. By considering the additional information of cross-sensitivity of air quality parameters, the performance of the calibration is enhanced substantially. Furthermore, the variances and means of different air quality measures are slightly dissimilar but their distributions are related to each other, as shown in Fig. \ref{boxer}. Particulate matter sensors have closely related data distributions among themselves. CO and $\textrm{SO}_2$ have a high degree of similarity too. These observations provide a rich evidence for utilizing their cross-sensitivities while designing the calibrator.   
The discrepancies in the low-cost LPNs result in missing values. The missing values are removed. In order to correct the measurements, a mean over an extended time is taken in \cite{badura2018evaluation}. Moreover, \cite{LOH2019452} presents a methodology using $1$ measurement per hour. We adopt a similar approach by averaging the data over past $1$ hour.  

\subsection{Validation - Stage X1}
\vspace{-1mm}
The past research on the calibration of the air quality parameter sensing mostly considers only single parameter, which is under study. We show that behavior in the form of X$1$ models for $\textrm{CO}, \textrm{ SO}_2, \textrm{ NO}_2, \textrm{ O}_3, \textrm{ PM}_{1.0}, \textrm{ PM}_{2.5} \textrm{ and} \textrm{ PM}_{10}$. In \cite{badura2018evaluation}, $\textrm{PM}_{2.5}$ is calibrated by developing only a $2$ dimensional space of calibrated and un-calibrated readings for a particular interval of time. Similarly, \cite {ferrer2020multisensor,barcelo2019self} propose methods for self calibration only. We develop the solution based upon that as a benchmark for comparison. Moreover, it is the only possible way to accomplish a reasonable estimation when only a single quantity is measured without access to environmental factors like temperature, pressure, and relative humidity. It is shown in Fig. \ref{rmse}. Except for $\textrm{PM}_{1.0}$ and $\textrm{CO}$, a fully connected deep neural network shows a superior calibration result. Since the dimensionality of the feature space is very limited, KNN and SVR also perform well. X$1$ is thus useful for feature deficient cases.

\vspace{-4mm}
\subsection{Validation - Stage X2}
Drawing inspiration from \cite{hagan2018calibration} and reproducing its results simply with temperature as an additional parameter would not have been a smart choice. Fig. \ref{corr} illustrates that not only does temperature have a stark correlation with air quality parameters, but the pressure and relative humidity also do. Therefore, for the calibration at stage X$2$, a strategy was employed to incorporate those environmental factors. It is enabled only when those factors are available to be utilized for making appropriate inferences. Interestingly, $\textrm{PM}_{10}$ has the least RMSE at the X$2$ stage when DNN was selected as the model. It can be explained through the behavior of the data under consideration where the $\textrm{PM}_{10}$ has the least cross-sensitivity. We implement the calibrator so that the output with the least RMSE is always returned as the final result. The cross-sensitivity has no role to play up to this stage. It serves the purpose of enhanced flexibility well, as it is likely for some cases where there exist only the environmental factors and not the other air quality measures. In such scenarios, the preceding stages shall be bypassed. 
\vspace{-4mm}
\subsection{Validation - Stage X3}
There is largely a void in the previous work on pivoting the cross-sensitivity of several air quality measures while calibrating the LPNs. In \cite{hagan2018calibration}, the researchers allude to this fact but only as a potential future work. We tend to develop a holistic approach with maximum information exploitation. The interdependence of the parameters is indicated in Fig. \ref{corr} as Pearson's correlation coefficients. Moreover, the datasheets of SPEC-sensors also provide analytical evidence of the cross-sensitivity of air quality measures.  Subfigures \ref{6a}, \ref{6b}, \ref{6c}, \ref{6d}, \ref{6e}, \ref{6f}, and \ref{6g} depict the improvement in the performance of calibrators and validate the performance of proposed methodology. The only anomaly is $\textrm{PM}_{10}$, where we observed a relatively low RMSE due to the low interdependence of $\textrm{PM}_{10}$ on the others.  Also, due to the existence of numerous air quality indicators, the model selection method searches for a larger space in this case. The number of combinations of input features can be $11+4 = 15$ because there are $4$ inputs corresponding to SPEC sensors and $3$ to PM sensors. Intuitively, fully connected tuned deep neural networks are the best learners in this space. The dimensionality of the data is high; consequently, more complex features can be learned \cite{lecun2015deep}. This approach, in conjunction with the control mechanism, epitomizes the modularity of the calibrator. Irrespective of which of the two or more air quality parameters are present, the proposed solution selects the best possible combination of inputs and the model. Since the model's selection is completed in the training phase, the size of the trained model is suitable to be deployed on-chip or in the server. In the worst case, the size DNN model is $2.5$ kB for $\textrm{NO}_2$, and for weights it is $823$ kB. So the total size is $825.5$ kB. For $7$ sensors $825.5\times11= 9080.5$ kB.  Such a model can easily be deployed on most edge devices including ESP$32$ (with $16$ MB EEPROM).

\vspace{-4mm}
\subsection{Validation - Stage X4}
The past work in air quality measures has shown keen interest in the interdependence of some of the gases and particulate matter. The research work presented in \cite{bartington2017patterns} and \cite{mccracken2013longitudinal} points out the relationship between $\textrm{PM}_{2.5}$ and CO. Stage X$4$ is driven by that motivation. $\textrm{PM}_{1.0}$ and $\textrm{CO}$ have shown a significant drop in RMSE. Whereas, for $\textrm{O}_3$, the RMSE experiences a marginal drop, as shown subfigures \ref{6b}, \ref{6e} and \ref{6f}. The bit $b_{n+m+5}$ bypasses the block when it is set to $0$; otherwise, it performs its operation depending upon the respective RMSE. Furthermore, the best performing models for the appropriate input combination results in encouraging results in the time domain. Fig. \ref{samples} presents the time samples representing the hours of a day. The proposed methodology maintains a high fidelity for all the air quality parameters. Specifically, the calibration results of the PM sensors are most effective as illustrated in subfigures \ref{7a}, \ref{7b} and \ref{7c}. The patterns are well learned due to lesser uncertainty. Moreover, the calibrated results of SPEC sensors are presented in subfigures \ref{7d}, \ref{7e}, \ref{7f} and \ref{7g}. 
Since the model has the capability to gather the output for each stage (X$1$, X$2$, X$3$, and X$4$), the stage with the best output is selected as the final calibrator, and the corresponding model parameters are transmitted to the edge devices via MQTT-broker.  
\vspace{-4mm}
\subsection{Validation - Forecasting}
Finally, the calibrated readings are fed to a separate agent inside the cloud integration server to select the relevant features based upon mutual information score. Those relevant features help to train a multivariate-LSTM model. Table \ref{Tab:for} provides the results in the form of RMSEs of the real and predicted values. Furthermore, a comparison with the approaches adopted in the past is also provided. It is evident from table \ref{Tab:for} that the $\textrm{SO}_2$ and $\textrm{NO}_2$ have not been reliably forecasted in the past as they have not yet been calibrated and predicted through modern machine learning tools. In conjunction with the LSTM core, our proposed feature selection methodology can enhance the accuracy many folds. The trend of predicted values for particulate matter are also dependable to a good extent as the $r^2$ values are above $80$ in all the cases.  
\begin{table}[!hbt]
\vspace{-4mm}
\caption{Air Quality Metric Forecast.}
	\centering
 \begin{tabular}{c | c  c | c }   

 \hline\hline
Air Quality Mertic& \multicolumn{2}{c|}{Proposed Architecture}& Past Work\\ [0.5ex] 
 \hline
Chemical Formula & RMSE& $r^2$&RMSE\\ [0.5ex] 
 \hline
 $\textrm{SO}_2$ & 0.02 & 0.96 &17.42\cite{li2018online} \\ 
\hline
CO & 0.004 & 0.75&0.342\cite{inproceedings1234}\\
\hline
$\textrm{NO}_2$ & 0.72 & 0.61&25.40\cite{li2018online} \\
\hline
$\textrm{O}_3$ & 0.050 & 0.92&2.36\cite{article} \\
\hline
$\textrm{PM}_{1.0}$ & 0.028 & 0.99&- \\
\hline
$\textrm{PM}_{2.5}$ & 0.061 & 0.96&0.092\cite{qi2018deep} \\
\hline
$\textrm{PM}_{10.0}$ & 2.66 & 0.81&26.8\cite{michaelides2017monitoring} \\
 \hline\hline
\end{tabular}
\label{Tab:for}
\vspace{-4mm}
\end{table}

\section{Conclusion}
We implement a flexible calibration and forecasting scheme for affordable smog monitoring, that effectively employs IoT architecture. The scheme provides measurement and calibration setup for smog-causing pollutants, namely, $\textrm{CO}$, $\textrm{SO}_2$, $\textrm{NO}_2$, $\textrm{O}_3$, $\textrm{PM}_{1.0}$, $\textrm{PM}_{2.5}$ and $\textrm{PM}_{10}$. We demonstrate that utilizing the cross-sensitivity of different air quality parameters during the later stages of our multi-stage calibration procedure allows flexible and affordable monitoring of smog constituents. To improve up the measurement accuracy of LPNs, we propose a flexible MAQ-CaF to enhance flexibility. The bidirectional and multi-topic capabilities of MQTT-broker are proposed for co-location calibration. MAQ-CaF can exploit tuned-machine learning techniques enabled through a systematically designed model selection methodology based upon the availability of data from specific low-cost sensors. The systematic validation results depict the reduced RMSE at the stages X$1$, X$2$, X$3$, and X$4$ for the pollutants under consideration. Moreover, the encouraging results of the forecast of air quality parameters for the future are determined by mutual information score-based feature selection method and multivariate-LSTM.  The technique can be effectively utilized in the developing world to improve their air quality tracking capability by leveraging IoT and flexible MAQ-CaF for detecting smog constituents in a wide geographic region. 
\vspace{-2mm}

\ifCLASSOPTIONcaptionsoff
  \newpage
\fi




\bibliographystyle{IEEEtran}
\bibliography{ref}

%

%





\end{document}